\documentclass[10pt,twocolumn,letterpaper]{article}

\usepackage[pagenumbers]{cvpr} 
\usepackage{graphicx}
\usepackage{amsmath}
\usepackage{amssymb}
\usepackage{booktabs}
\usepackage{color}
\usepackage{colortbl}  
\usepackage{xcolor}
\usepackage{array}
\usepackage{bbding}
\definecolor{hollywoodcerise}{rgb}{0.96, 0.0, 0.63}
\definecolor{lasallegreen}{rgb}{0.03, 0.47, 0.19}
\definecolor{hanpurple}{rgb}{0.32, 0.09, 0.98}
\definecolor{green(pigment)}{rgb}{0.0, 0.65, 0.31}
\usepackage{multirow}
\newcommand{\aka}{\textit{a.k.a.}}

\definecolor{cvprblue}{rgb}{0.21,0.49,0.74}
\usepackage[pagebackref,breaklinks,colorlinks,citecolor=cvprblue]{hyperref}

\def\paperID{6652} 
\def\confName{CVPR}
\def\confYear{2024}

\title{UniBind: LLM-Augmented Unified and Balanced Representation Space to Bind Them All}
\author{Yuanhuiyi Lyu$^{1}$ $^{*}$ \quad Xu Zheng$^{1}$ \thanks{Equal Contribution.} \quad Jizhou Zhou$^{1}$ \quad Lin Wang$^{1}$$^{,2}$ \thanks{Corresponding author.}\\
$^{1}$AI Thrust, HKUST(GZ) \quad $^{2}$Dept. of CSE, HKUST 
\\
{\tt\small \{yuanhuiyilv, jiazhouzhou\}@hkust-gz.edu.cn, zhengxu128@gmail.com, linwang@ust.hk}
\\
\small{Project Page: \url{https://vlislab22.github.io/UniBind/}}
}

\begin{document}
\maketitle



\begin{abstract}
We present \textbf{UniBind}, a flexible and efficient approach that learns a unified representation space for seven diverse modalities -- image, text, audio, point cloud, thermal, video, and event data. 
Existing works, \eg, ImageBind~\cite{girdhar2023imagebind}, 
treat the image as the central modality and build an image-centered representation space; however, the space may be sub-optimal as it leads to an unbalanced representation space among all modalities.  
Moreover, the category names are directly used to extract text embeddings for the downstream tasks, making it hardly possible to represent the semantics of multi-modal data.
The `out-of-the-box' insight of our UniBind is to make the alignment centers modality-agnostic and further learn a unified and balanced representation space, empowered by the large language models (LLMs). 
UniBind is superior in its flexible application to all CLIP-style models and delivers remarkable performance boosts. 
To make this possible, we 1) construct a knowledge base of text with the help of LLMs and multi-modal LLMs;
2) adaptively build LLM-augmented class-wise embedding centers on top of the knowledge base and encoded visual embeddings;
3) align all the embeddings to the LLM-augmented embedding centers via contrastive learning to achieve a unified and balanced representation space.
UniBind shows strong zero-shot recognition performance gains over prior arts by an average of 6.36\%.
Finally, we achieve new state-of-the-art performance, \eg, a 6.75\% gain on ImageNet, on the multi-modal fine-tuning setting while reducing 90\% of the learnable parameters. 
\end{abstract}

\begin{figure}[t!]
    \centering
    \includegraphics[width=\linewidth]{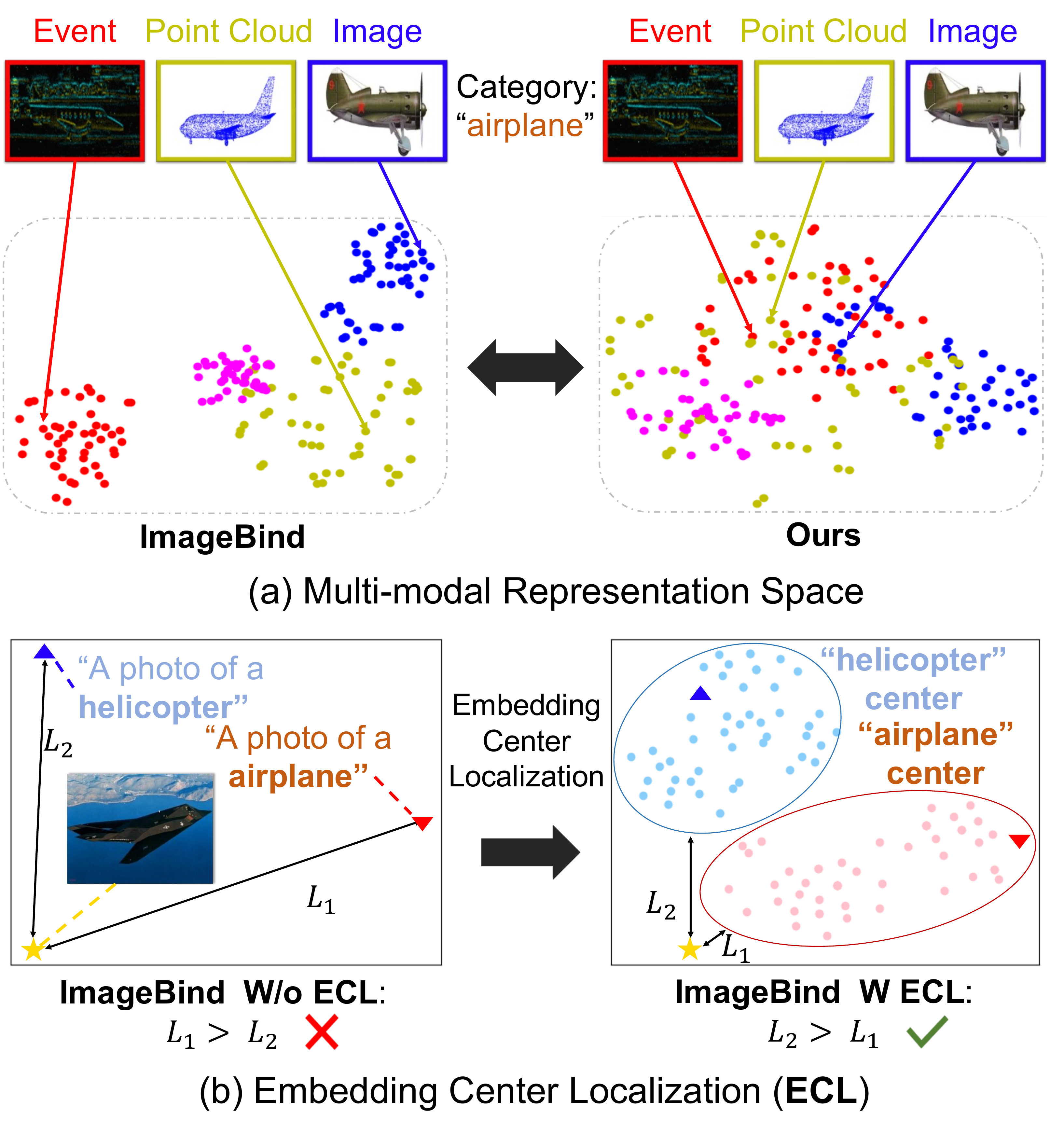}
    \vspace{-20pt}
    \caption{(a) By making the alignment center modality-agnostic, our UniBind can learn a unified and balanced representation space. 
    (b) The embedding centers for each semantic category: these centers exhibit more complementary semantics compared to embeddings solely encoded by category names.}
    \label{tease_fig}
\end{figure}
\section{Introduction}
\label{sec:intro}
Humans use multiple senses-- each of which is from a different source, \aka, modality-- to perceive and interpret the world~\cite{ngiam2011multimodal,su2023recent}.
Humans are naturally equipped with the capacity to process and fuse multiple modalities simultaneously. 
For machines to emulate human intelligence, it is imperative for them to interpret, reason, and fuse multi-modal inputs, such as vision, text, audio, \etc~\cite{duan2022multimodal}.
This has inspired many methods that employ paired data 
to align image with text~\cite{frome2013devise, su2019vl, chen2020uniter, li2020oscar} or align image with audio~\cite{guzhov2022audioclip, mahmud2023ave}.
Building on these works, early research has largely focused on integrating additional modalities, \eg, CLIP2Video~\cite{fang2021clip2video} and PointCLIP~\cite{zhang2022pointclip} for enhancing the comprehensiveness and accuracy of multi-modal data representation, and ultimately, improving performance across various tasks.
Recent endeavors have shown the possibility of learning across multiple modalities, 
including video~\cite{sun2023learning, cheng2023cico, zheng2023cvt}, point cloud~\cite{zhang2022pointclip, zhu2022pointclip, huang2022clip2point, wang2022multimodal}, thermal~\cite{girdhar2023imagebind, zhang2023cmx}, event~\cite{zhou2023clip, zheng2023deep}, \etc. 
Among these methods, ImageBind ~\cite{girdhar2023imagebind} sets a new way to learn a single shared representation space by leveraging multiple types of image-paired data. It utilizes the binding property of image modality to align the embeddings from the other modalities with the image embeddings. 

However, as depicted in Fig.~\ref{tease_fig} (a), treating the image as the central modality and building an image-centered representation space leads to sub-optimal results, it may introduce bias and thus results in an unbalanced representation space among all modalities~\cite{el2023learning}. 
Also, as depicted in Fig.~\ref{tease_fig} (b), existing CLIP-style models, \eg, ImageBind~\cite{girdhar2023imagebind}, solely utilize the text embeddings obtained from category names as embedding centers for categories. Nonetheless, category names, such as {\fontfamily{qcr}\selectfont [‘Airplane']} and {\fontfamily{qcr}\selectfont [‘Helicopter']}, may not fully represent the semantics of the visual data, as there exist numerous images of airplanes with varying backgrounds and conditions.

This paper strives to tackle two problems: 1) the unbalanced representation space resulting from taking a specific visual modality as the alignment center, and 2) the unreliable nature of embedding alignment centers that rely solely on category names. Accordingly,
we propose UniBind, a flexible and efficient approach for binding seven modalities-- image, text, audio, point cloud, thermal, video, and event data. The core insight of our UniBind is to make the alignment centers modality-agnostic and further learn a unified and balanced representation space, empowered by the large language models (LLMs) and multi-modal large language models (multi-modal LLMs). Our UniBind is superior in its flexible application to all CLIP-style models and delivers remarkable performance boosts (+3.83\% in N-caltech~\cite{orchard2015converting} with E-CLIP~\cite{zhou2023clip}).

Specifically, we first construct a knowledge base of texts which are extracted from the text generated by several LLMs, \eg, GPT-4~\cite{openai2023gpt4} and LLaMa~\cite{touvron2023llama}, as well as multi-modal LLMs, \eg, BLIP-2~\cite{li2023blip} and LLaMa-Adapter~\cite{zhang2023llama}. In practice, GPT-4 and LLaMa are utilized to generate the category descriptions, while BLIP-2 and LLaMa-Adapter are used to provide the multi-modal data descriptions (Sec.~\ref{sec:knowledgebase}). 
Secondly, we compute the class-wise similarity between the input prompts and the text embeddings. It then utilizes the top 50 text embeddings to construct the corresponding class-wise text embedding center (Sec.~\ref{sec:ECL}). For example, as depicted in Fig.~\ref{tease_fig} (b), we select the top 50 text embeddings based on their similarity to the input prompts: {\fontfamily{qcr}\selectfont ["A photo of helicopter/airplane."]}. These selected embeddings are then utilized to construct the embedding centers for the categories of {\fontfamily{qcr}\selectfont [`helicopter']} and {\fontfamily{qcr}\selectfont [`airplane']}.
Lastly, we align all modality embeddings toward the text embedding centers using contrastive learning loss functions (Sec.~\ref{sec:unifiedlearning}). This ensures that all modalities are equally considered in the representation space and achieve a unified and balanced representation space, as shown in Fig.~\ref{tease_fig} (a). 

We apply our UniBind to the state-of-the-art (SoTA) CLIP-style multi-modal learning methods, including CLIP~\cite{radford2021learning}, E-CLIP~\cite{zhou2023clip}, Audio-CLIP~\cite{guzhov2022audioclip}, Point-CLIP~\cite{zhang2022pointclip}, ImageBind~\cite{girdhar2023imagebind}, and PointBind~\cite{guo2023point}, on 14 benchmarks from seven modalities. Note that, our UniBind is the first work to introduce the event modality~\cite{zheng2023deep} into the multi-modal representation space. UniBind consistently delivers significant performance improvements with all the CLIP-style multi-modal methods on all the benchmarks from the seven modalities, such as +5.55\% in ImageNet-1K~\cite{deng2009imagenet} with ImageBind~\cite{girdhar2023imagebind} and +8.28\% in N-caltech~\cite{orchard2015converting} with PointBind~\cite{guo2023point}. 
Moreover, we achieve new SoTA performance, \eg, +6.75\% gain on ImageNet-1K with the multi-modal fine-tuning setting while reducing 90\% of the learnable parameters.
Additionally, in the cross-modal retrieval tasks, our UniBind demonstrates a substantial performance improvement by +17.96\% with PointBind on the top-20 recall score in the event-to-image retrieval task.

\section{Related Work}
\noindent \textbf{Multi-modal Learning:}
From the modality alignment perspective, existing methods can be divided into two categories: alignment at the token and feature levels.
(1) Token-level alignment methods~\cite{zhang2023meta, fang2021clip2video, mukhoti2023open, zheng2023cvt, cheng2023cico, zhang2021vinvl} align multi-modal token embeddings in a shared token embeddings space and design a subsequent encoder to extract the feature of these input token embeddings. 
(2) Feature-level alignment methods are based on the unified vision-language representation space, built by the CLIP style large vision-language models~\cite{radford2021learning, li2022blip, wei2023iclip, liu2022universal, luo2022clip4clip, xue2022clip, lyu2024image}, \eg, BLIP~\cite{li2022blip}. These methods adapt one~\cite{zhou2023clip,zhang2022pointclip, zhu2022pointclip,guzhov2022audioclip,mahmud2023ave,fang2021clip2video} or more~\cite{girdhar2023imagebind, guo2023point} modalities to the image representation space to align multiple visual modalities.
Representative works include ImageBind~\cite{girdhar2023imagebind}, which learns a single shared representation space by leveraging multiple types of image-paired data. It leverages the binding property of images and aligns each modality's embeddings to image embeddings. Other works, such as PointCLIP~\cite{zhang2022pointclip, zhu2022pointclip} and AudioCLIP~\cite{guzhov2022audioclip}, align point cloud and audio modalities, respectively, to the image representation space using cross-modal correlation or attention mechanisms.
However, since these methods treat the image modality as the multi-modal alignment center, the obtained representation space is unbalanced among all the visual modalities~\cite{el2023learning}, as demonstrated in Fig.~\ref{tease_fig}.
By contrast, we propose to learn modality-agnostic alignment centers, buttressed by the LLMs, thus yielding a unified and balanced visual representation space. 
Our UniBand binds the multi-modalities with the same semantics to bridge the gap of multi-modalities.

\begin{figure*}[t!]
    \centering
    \includegraphics[width=\textwidth]{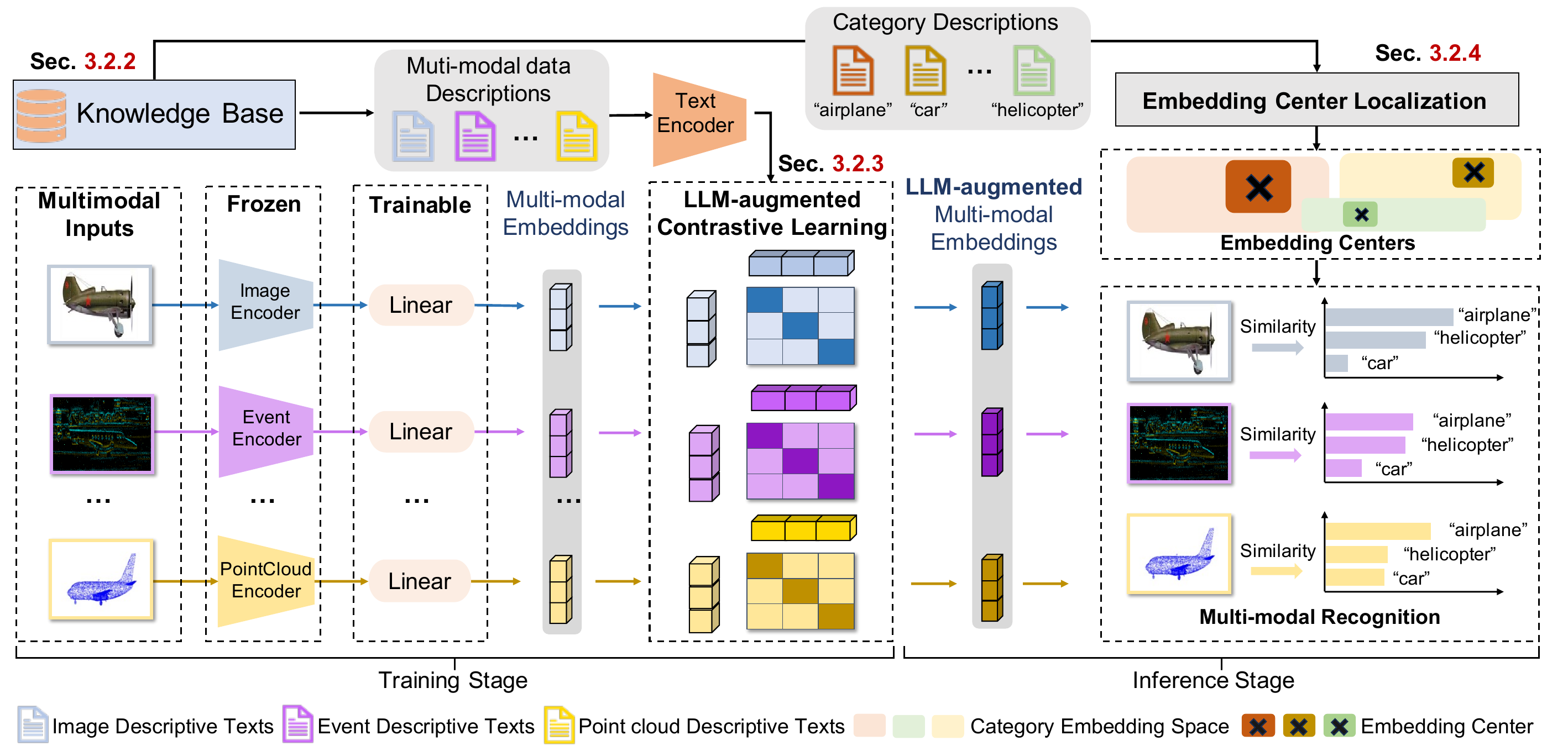}
    \vspace{-18pt}
    \caption{An overview of our UniBind. Firstly, we construct the knowledge base and then learn a unified representation space via LLM-augmented contrastive learning. Lastly, We utilize the embedding center localized by the knowledge base to obtain the predictions.}
    \label{overall}
\end{figure*}

\noindent \textbf{LLMs and Knowledge Base:} 
In Natural Language Processing (NLP), knowledge bases are widely used to enhance the understanding of human language~\cite{hu2021knowledgeable, alkhamissi2022review, das2022knowledge, sen2022logical} and the robustness of generated results~\cite{gui2021kat, yang2022enhancing, salemi2023pre}. With the development of LLMs~\cite{touvron2023llama, zhao2023survey}, researchers use them to build knowledge bases~\cite{ma2023query, li2023flexkbqa, carta2023iterative, nayak2023llm2kb}. These methods can be categorized into two groups,
including: \textbf{1)} designing effective text prompts via LLMs to enhance the representation ability of the text encoder~\cite{santu2023teler, lian2023llm}; and \textbf{2)} utilizing LLMs to verbalize the semantics of text input by generating texts with similar semantics for the same category data, aiming to enhance the robustness of text representation~\cite{zhu2022pointclip, naeem2023i2mvformer}. 
We utilize LLMs and multi-modal LLMs to construct a knowledge base to incorporate the prior knowledge of each category by generating the descriptions of the category name. This augments the text embeddings of the category names in the representation space.

\noindent \textbf{Language-augmented Representation Learning} aims to enhance the visual representation space by incorporating language, \ie, text data. As pointed out by~\cite{el2023learning}, language helps to identify conceptually similar image pairs even if they are visually dissimilar in visual recognition tasks.
Moreover, efforts have been taken to leverage text as the representation centers for contrastive learning in information retrieval~\cite{liu2022universal, yang2022unified}. For example, UniVL-DR~\cite{liu2022universal} addresses the modality gap by verbalizing images to text and constructs a unified representation space for multi-modal dense retrieval, resulting in significant performance gains. 
Differently, we introduce a text embedding center strategy to the multi-modal domain. Our UniBind utilizes extracted text embeddings as the alignment centers and further binds the visual modality embeddings, thereby facilitating multi-modal representation learning and obtaining a balanced and unified representation space.

\vspace{-8pt}
\section{The Proposed UniBind}

\subsection{Problem Setting and Overview}
\noindent \textbf{Problem setting:} 
We follow the multi-modal recognition setting popularized by ImageBind~\cite{girdhar2023imagebind}. It uses the default set of text prompt templates ${P_j^1, P_j^2, ..., P_j^n}$ from CLIP~\cite{radford2021learning}. It then computes the similarity score between the input multi-modal data $V_i$ and the $C_j$ category by:
\begin{small}
\begin{equation}
\label{sourcesimliraty}
    S_{(V_i,C_j)} = cos<F^{V}(V_i), mean\{ F^{T}({P_j^1, ..., P_j^n})\}>,
\end{equation}
\end{small}
where $F^{V}$ and $F^{T}$ represent the visual and text encoders, respectively, which extract the image and text embeddings.
The key insight of our UniBind is to make the alignment centers modality-agnostic and then learn a unified and balanced representation space for diverse modalities by leveraging the embedding centers constructed from the knowledge base, thereby binding them together. UniBind strives to address the two main challenges: 
1) The unbalanced representation space that emerges from designating a particular visual modality as the alignment center, and 2) The unreliable nature of embedding alignment centers that exclusively depend on category names.
To this end, firstly, UniBind constructs a knowledge base of text embeddings using LLMs and Multi-modal LLMs. Secondly, UniBind~\cite{girdhar2023imagebind} adaptively builds LLM-augmented class-wise embedding centers based on the knowledge base and aligns multi-modal embeddings to the embedding centers with contrastive learning to build a unified embedding space.

\noindent \textbf{Overview:} 
An overview of UniBind is shown in Fig.~\ref{overall}. 
Specifically, given $n$ visual modalities, Our UniBind includes $n$ multi-modal encoders $F_n$ and a text encoder ${F^T}$, which are adapted from existing multi-modal learning models such as ImageBind~\cite{girdhar2023imagebind}. The only modification of these multi-modal learning models made in UniBind is adding a trainable linear layer to each of the $n$ visual encoders, while the adopted encoders are all frozen during training. Our framework includes two stages: \\
1) Training Stage:
For training, we initially construct the knowledge base (Sec.\ref{sec:knowledgebase}) by incorporating both LLMs and multi-modal LLMs. We then leverage the knowledge base to learn a unified representation space (Sec.\ref{sec:unifiedlearning}) via LLM-augmented contrastive learning. \\
2) Inference Stage:
Building on our unified multi-modal representation space, we infer recognition results via a novel Embedding Center Localization module (Sec.~\ref{sec:ECL}). We now describe these technical components in detail.

\begin{figure}[t!]
    \centering
    \includegraphics[width=\linewidth]{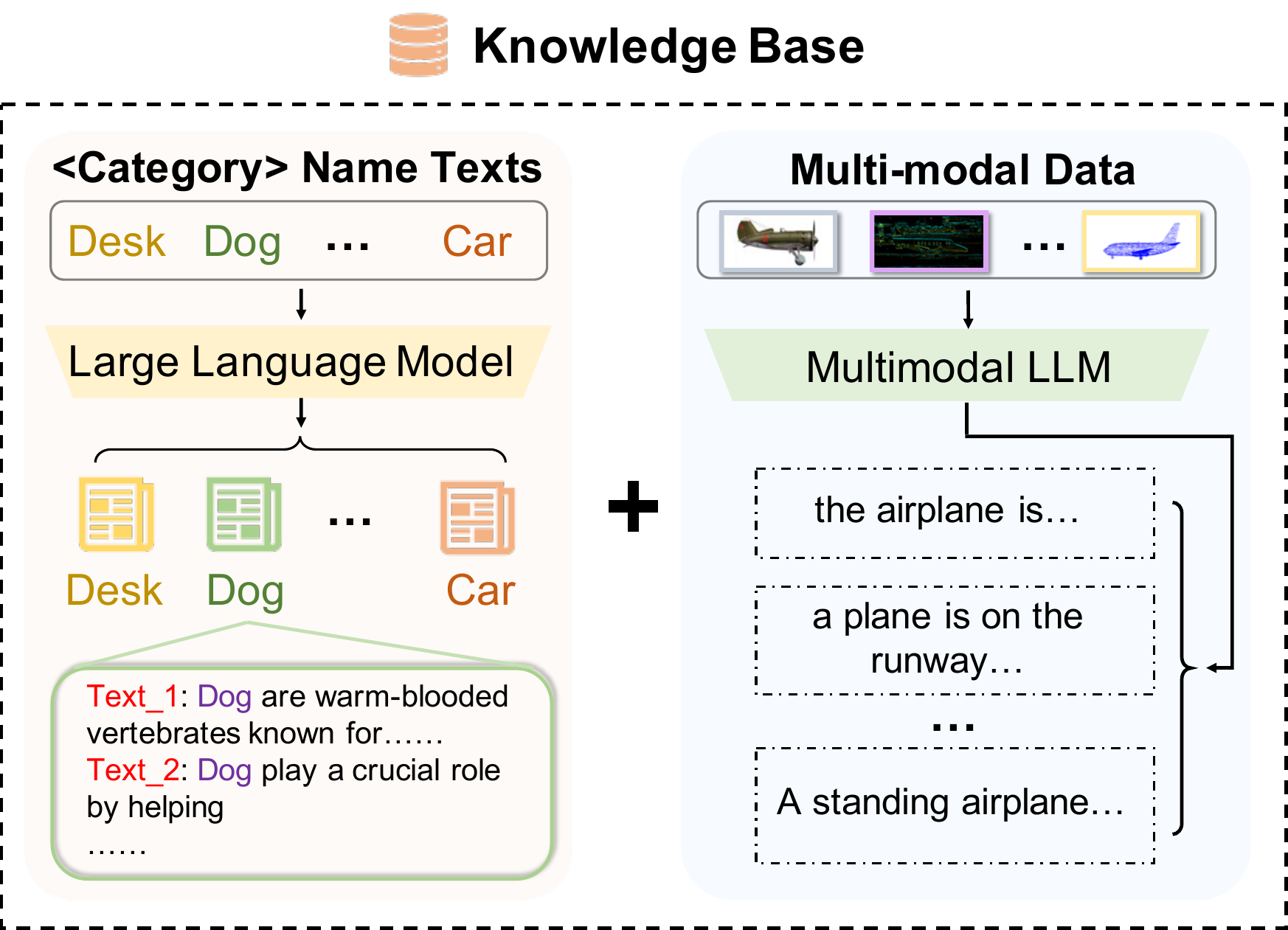}
    \vspace{-18pt}
    \caption{Knowledge Base. Generation pipeline for category descriptions (left) and multi-modal data descriptions (right).}
    \label{K_database}
\end{figure}

\vspace{-7pt}
\subsection{Knowledge Base Construction}
\label{sec:knowledgebase}
The construction of the knowledge base comprises two parts: 1) Category descriptions, generated by LLMs. These texts are employed to localize the embedding centers. 2) Multi-modal data descriptions produced by multi-modal LLMs. These descriptions help to alleviate the modality gaps prevalent among multiple modalities.

Although it has been demonstrated in~\cite{el2023learning, liu2022universal} that language is a powerful tool for capturing semantic relationships among multi-modal data, dependency on category names exclusively to align modalities with extracted embeddings is unreliable. As discussed in Sec.~\ref{sec:intro}, category names cannot fully capture the semantics of multi-modal data. To address this issue, we first use LLMs, such as GPT-4~\cite{openai2023gpt4} and LLaMa~\cite{touvron2023llama}, to generate category descriptions based on the category names, as shown in Fig.~\ref{K_database}: 
\begin{equation}
     T_{C_i}^1, ..., T_{C_i}^n = F^{LLMs}(C_i),
\end{equation}
where the $T_{C_i}^1, ..., T_{C_i}^n$ are the $n$ generated descriptions for the category $C_i$ and $\mathcal{F}^{LLMs}$ are the aforementioned LLMs. 
Subsequently, we produce descriptions for multi-modal data via multi-modal LLMs:
\begin{equation}
    T_{I_i}, ..., T_{A_i} = F^{MLLMs}(I_i), ..., F^{MLLMs}(A_i),
\end{equation}
here, $I_i, ..., A_i$ represent the multi-modal data inputs, while $T_{I_i}, ..., T_{A_i}$ are the generated multi-modal data descriptions
and $F^{MLLMs}$ denote the aforementioned multi-modal LLMs.
As an example, consider the category {\fontfamily{qcr}\selectfont [`Desk']}, we use LLMs to generate category descriptions, such as {\fontfamily{qcr}\selectfont ["A computer monitor is prominently displayed on the desk, indicating it is a workstation"]}. We then utilize multi-modal LLMs to generate descriptions for image data falling within the {\fontfamily{qcr}\selectfont [`Desk']} category. Finally, we compile these two sets of descriptions to formulate our knowledge base.

\subsection{Unified Representation Space Learning}
\label{sec:unifiedlearning}
Expanding upon our knowledge base, we subsequently align multiple modalities to learn a unified multi-modal representation space.
For multi-modal data, we utilize feature encoders derived from existing multi-modal models with frozen parameters and learnable subsequent linear layers, to obtain embeddings for each modality:
\begin{equation}
    v_{I_i}, ..., v_{A_i} = F_{I}(I_i), ..., F_{A}(A_i),
\end{equation}
where the $v_{I_i}, ..., v_{A_i}$ are the extracted embeddings and $F_{I}(\cdot), ..., F_{A}(\cdot)$ are the feature encoders for each modality.
Subsequently, we generate text embeddings of multi-modal data descriptions $T_{I_i}, ..., T_{A_i}$ via text encoder:
\begin{equation}
    z_{I_i}, ..., z_{A_i} = F^T(T_{I_i}), ..., F^T(T_{A_i}),
\end{equation}
where the $T_{I_i}, ..., T_{A_i}$ are the generated multi-modal data descriptions, $F^{T}(\cdot)$ is the text encoder, and $z_{I}, ..., z_{A}$ are the extracted text embeddings. 
The extracted visual and text embeddings are employed for learning a unified representation space. In contrast to existing multi-modal learning frameworks, such as ImageBind~\cite{girdhar2023imagebind}, we do not impose contrastive learning objectives among visual data with the image center. 
Instead, we impose contrastive learning objectives directly between the multi-modal and text embeddings. 
As an illustration, for aligning the visual modality $I$ to our unified representation space, the extracted visual embeddings ${v_{I_1}, ..., v_{I_n}}$ and the corresponding text embeddings ${z_{I_1}, ..., z_{I_n}}$ are employed for contrastive learning within this representation space. Text embeddings generated from corresponding descriptions are considered positive samples for input visual data, whereas text embeddings from other visual data are utilized as negative samples:
\begin{small}
\begin{equation}
    \mathcal{L}_{(\mathcal{I},\mathcal{A})} = -log\frac{exp(v_{I_i}^T \cdot z_{I_i})/\tau}{exp(v_{I_i}^T \cdot z_{I_i}/\tau)+\sum_{j \ne i}(v_{I_i}^T \cdot z_{I_j}/\tau)},
\end{equation}
\end{small}
where $z_{I_j}$ is the corresponding text embeddings of the visual data $I_j$ in visual modality $I$.

\begin{figure}[t!]
    \centering
    \includegraphics[width=0.96\linewidth]{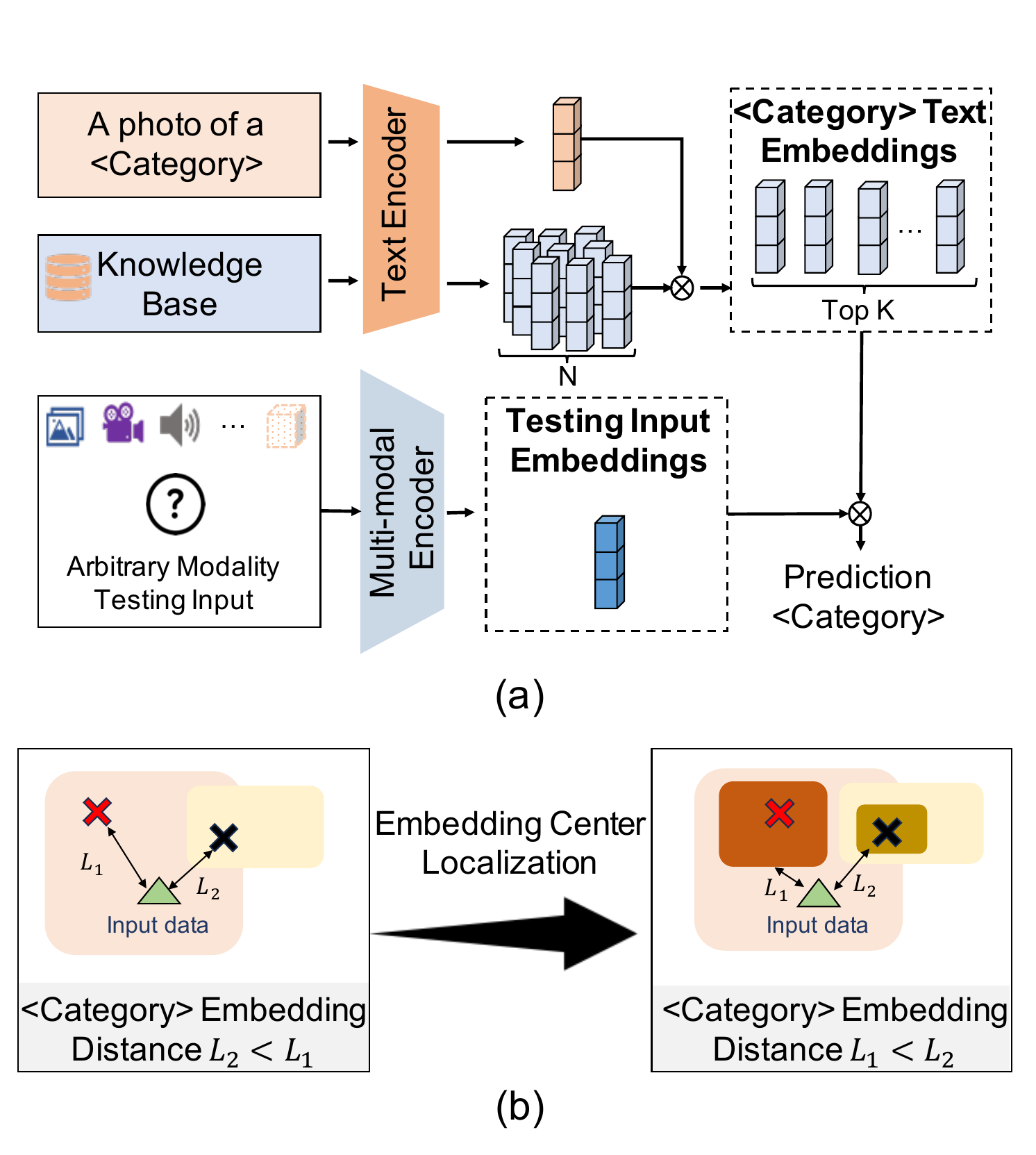}
    \vspace{-12pt}
    \caption{(a) The details for our embedding center localization. (b) The impact of our embedding center localization is demonstrated.}
    \label{localization}
\end{figure}
\subsection{Embedding Center Localization}
\label{sec:ECL}
We localize our embedding centers by selecting 50 text embeddings for each category from the knowledge base. Specifically, we utilize the basic prompt {\fontfamily{qcr}\selectfont ["A photo of a [Category]"]} to calculate cosine similarity, subsequently selecting the top 50 descriptions based on the highest cosine similarity scores. As depicted in Fig.~\ref{localization} (a), we subsequently derive the text embedding center $EC_i$ for category $C_i$ from these top 50 descriptions by processing them through the text encoder $F^T$:
\begin{equation}
    EC_i = \{z_{C_i}^1, ..., z_{C_i}^{50}\} = F^T(T_{C_i}^1), ..., F^T(T_{C_i}^{50}),
\end{equation}
The text embeddings generated from the top 50 descriptions collectively form the embedding center $EC_i$ for each category. Consequently, UniBind extracts embedding centers from the knowledge base which have more complementary semantics than simply using the category names. Unlike existing methods exemplified in Eq.~\ref{sourcesimliraty}, our embedding centers \textit{establish more distinct category boundaries in representation space} (shown in Fig.~\ref{ecl_ab}). For instance, we compute the similarity between an arbitrary modality input $M_i$ and the category $C_j$ as follows:
\begin{equation}
\label{unisimilarity}
    S_{(M_i,EC_j)} = max\{cos<F_{m}(M_i), z_{C_i}^1, ..., z_{C_i}^{50}>\},
\end{equation}
where the $z_{I}, ..., z_{A}$ are the extracted text embeddings of category $C_i$. As shown in Fig.\ref{localization} (b), compared with the mean of text prompt embeddings, our embedding centers have more significant spatial distributions. With our embedding centers, multi-modal data distributed at the boundary of the category representation space can effectively avoid interference from other categories (shown in Fig.~\ref{ecl_ab}), thus facilitating more accurate recognition.

\begin{table}[t!]
\renewcommand{\tabcolsep}{1pt}
\resizebox{\linewidth}{!}{
\begin{tabular}{llccc}
\toprule
\multicolumn{1}{c}{Modalities} & \multicolumn{1}{c}{Dataset} & \multicolumn{1}{c}{Metric} & \multicolumn{1}{c}{Scale} & \multicolumn{1}{c}{\#cls}\\ \midrule
\multirow{3}{*}{\textcolor{blue!70!black}{Image}} & \textcolor{blue!70!black}{ImageNet-1K (IN-1K)}~\cite{deng2009imagenet} & Acc & 1,280K & 1,000\\ 
& \textcolor{blue!70!black}{Places-Stanford-365 (P365)} ~\cite{lopez2020semantic}  & Acc & 1,240K & 365\\
& \textcolor{blue!70!black}{Caltech-101 (cal)} ~\cite{fei2004learning}  & Recall & 8K & 101\\ \midrule
\multirow{2}{*}{\textcolor{yellow!70!black}{PointCloud}} &\textcolor{yellow!70!black} {ModelNet-40 (ModelNet40)}~\cite{wu20153d}  & Acc & 9K & 40\\
& \textcolor{yellow!70!black}{ShapeNet-part (ShapeNet)}~\cite{chang2015shapenet}  & Acc &  16K & 16\\ \midrule
\multirow{2}{*}{\textcolor{magenta!70!black}{Audio}} & \textcolor{magenta!70!black}{ESC 5-folds (ESC)}~\cite{piczak2015esc}  & Acc & 2K & 50 \\
& \textcolor{magenta!70!black}{Urban-Sound-8K (Urban-S)}~\cite{salamon2014dataset} & Acc & 8K & 10 \\ \midrule
\multirow{2}{*}{\textcolor{brown}{Thermal}} & \textcolor{brown}{LLVIP (LLVIP)} ~\cite{jia2021llvip} & Acc & 15K & 2\\
& \textcolor{brown}{RGB-T Selected (RGB-T)} ~\cite{hwang2015multispectral} & Acc & 10K & 2\\ \midrule
\multirow{2}{*}{\textcolor{violet}{Video}} & \textcolor{violet}{MSR-VTT (MSR-VTT)}~\cite{xu2016msr}  & Acc  & 10K & 20\\
& \textcolor{violet}{UCF-101 (UCF-101)}~\cite{soomro2012ucf101} & Acc & 14K & 101\\ \midrule
\multirow{2}{*}{\textcolor{red!70!black}{Event}} & \textcolor{red!70!black}{N-Caltech-101 (N-cal)}~\cite{orchard2015converting} & Acc \& Recall & 8K & 101\\
& \textcolor{red!70!black}{N-ImageNet-1K (N-IN-1K)}~\cite{kim2021n}& Acc & 1,280K & 1,000\\
\bottomrule
\end{tabular}}
\caption{Summary of experimental settings across various modalities. We report the task, dataset, and data scale for each modality.}
\label{Table_dataset_1}
\end{table}

\begin{table*}[t!]
\renewcommand{\tabcolsep}{2pt}
\resizebox{\linewidth}{!}{
\begin{tabular}{lcccccccccccc}
\toprule
\multicolumn{1}{c|}{Model} & \multicolumn{2}{c|}{\textcolor{blue!70!black}{Image}} & \multicolumn{2}{c|}{\textcolor{yellow!70!black}{Point Cloud}} & \multicolumn{2}{c|}{\textcolor{magenta!70!black}{Audio}} & \multicolumn{2}{c|}{\textcolor{brown}{Thermal}} & \multicolumn{2}{c|}{\textcolor{violet}{Video}} & \multicolumn{2}{c}{\textcolor{red!70!black}{Event}} \\ \cmidrule{2-13}
\multicolumn{1}{l|}{} & \multicolumn{1}{c|}{\textcolor{blue!70!black}{IN-1K}} & \multicolumn{1}{c|}{\textcolor{blue!70!black}{Place-365}} & \multicolumn{1}{c|}{\textcolor{yellow!70!black}{ModalNet40}} & \multicolumn{1}{c|}{\textcolor{yellow!70!black}{ShapeNet}} & \multicolumn{1}{c|}{\textcolor{magenta!70!black}{ESC-50}} & \multicolumn{1}{c|}{\textcolor{magenta!70!black}{Urban-S}} & \multicolumn{1}{c|}{\textcolor{brown}{LLVIP}} & \multicolumn{1}{c|}{\textcolor{brown}{RGB-T}} & \multicolumn{1}{c|}{\textcolor{violet}{MSR-VTT}} & \multicolumn{1}{c|}{\textcolor{violet}{UCF-101}} & \multicolumn{1}{c|}{\textcolor{red!70!black}{N-Cal}} & \multicolumn{1}{c}{\textcolor{red!70!black}{N-IN-1K}} \\ \midrule
\multicolumn{13}{c}{\textbf{Fine-tuning Setting}} \\ \midrule
Meta-Transformer~\cite{zhang2023meta} & 83.10 & 52.70 & 90.50 & 99.30 & - & - & - & - & - & 46.60 & \tiny{\XSolidBrush} & \tiny{\XSolidBrush} \\
ImageBind~\cite{girdhar2023imagebind} w/ linear & 80.19 & 49.45 & \tiny{\XSolidBrush} & \tiny{\XSolidBrush} & 83.40 & \textbf{71.60} & - & 60.55 & 63.81 & \textbf{98.06} & \tiny{\XSolidBrush} & \tiny{\XSolidBrush} \\
PointBind~\cite{guo2023point}  w/ linear & 80.19 & 49.45 & 90.64 & 99.09 & 83.40 & \textbf{71.60} & - & 60.55 & 63.81 & \textbf{98.06} & \tiny{\XSolidBrush} & \tiny{\XSolidBrush} \\
PointBind (\textbf{+Event}) & 80.19 & 49.45 & 90.64 & 99.09 & 83.40 & \textbf{71.60} & - & 60.55 & 63.81 & \textbf{98.06} & 77.83 &  23.69\\
\textbf{Ours w/ PointBind} & \textbf{86.94} & \textbf{56.99} & \textbf{90.72} & \textbf{99.59} & \textbf{84.01} & 69.09 & - & \textbf{60.71} & \textbf{69.53} & 93.31 & \textbf{78.05} & \textbf{24.48} \\
$\Delta $ & {\textbf{+6.75}} & {\textbf{+7.54}} & {\textbf{+0.08}} & {\textbf{+0.50}} & {\textbf{+0.61}} & -2.51 & - & {\textbf{+0.16}} & {\textbf{+5.72}} & -4.75 & {\textbf{+0.22}} & {\textbf{+0.79}}
\\ \midrule
\multicolumn{13}{c}{\textbf{Zero-shot Setting}} \\ \midrule

ImageBind~\cite{girdhar2023imagebind} & 77.70 & 45.40 & \tiny{\XSolidBrush} & \tiny{\XSolidBrush} & 66.90 & 41.73 & 63.40 & 54.71 & 31.27 & 64.84 & \tiny{\XSolidBrush} & \tiny{\XSolidBrush} \\
PointBind~\cite{guo2023point} & 77.70 & 45.40 & 77.67 & 98.85 & 66.90 & 41.73 & 63.40 & 54.71 & 31.27 & 64.84 & \tiny{\XSolidBrush} & \tiny{\XSolidBrush} \\
PointBind (\textbf{+Event}) & 77.70 & 45.40 & 77.67 & 98.85 & 66.90 & 41.73 & 63.40 & 54.71 & 31.27 & 64.84 & 50.98 & 10.79 \\
\textbf{Ours w/ PointBind} & \textbf{83.25} & \textbf{53.84} & \textbf{80.59} & \textbf{98.96} & \textbf{71.70} & 62.56 & \textbf{64.67} & \textbf{56.20} & \textbf{40.90} & \textbf{73.74} & \textbf{59.26} & \textbf{13.85} \\
$\Delta $ & {\textbf{+5.55}} & {\textbf{+8.44}} & {\textbf{+2.92}} & {\textbf{+0.11}} & {\textbf{+4.80}} & {\textbf{+20.83}} & {\textbf{+1.27}} & {\textbf{+1.49}} & {\textbf{+9.63}} & {\textbf{+8.90}} & {\textbf{+8.28}} & {\textbf{+3.06}} \\ \bottomrule
\end{tabular}}
\caption{Emergent zero-shot and fine-tuning recognition on six modalities. 
}
\label{Table_main_1}
\end{table*}
\subsection{Implementation}
UniBind can be flexibly implemented with different existing CLIP-style multi-modal learning models, such as PointCLIP~\cite{zhang2022pointclip}, ImageBind~\cite{girdhar2023imagebind} and PointBind~\cite{guo2023point}. 

\noindent \textbf{Backbone Models} We use existing CLIP style multi-modal learning models as the backbones to implement our UniBind. Concretely, we implement our UniBind with the following models: CLIP~\cite{radford2021learning}, ImageBind~\cite{girdhar2023imagebind}, PointBind~\cite{guo2023point}, E-CLIP~\cite{zhou2023clip}, PointCLIP~\cite{zhu2022pointclip}, and AudioCLIP~\cite{guzhov2022audioclip}. 
We use separate visual encoders for image, point cloud, audio, thermal, video, and event data. 
We add a simple linear layer at the end of the visual encoders of each modality for mapping the multi-modal embeddings to our unified representation space. 
We utilize the frozen text encoder from the backbone model as our text encoder. 

\noindent \textbf{Training and Inference} UniBind can be utilized for both zero-shot and fine-tuning recognition tasks. For zero-shot tasks, as depicted in Fig.~\ref{overall}, UniBind employs the basic prompt {\fontfamily{qcr}\selectfont ["A photo of a [Category]"]} to select the top 50 most related text embeddings from the knowledge base as class-wise embedding centers. The similarities between the multi-modal embeddings and the class-wise embedding centers are then utilized to make recognition predictions. For fine-tuning, there is a training stage with the proposed representation space learning, as shown in Fig.~\ref{overall}. The inference process is the same as zero-shot.

\vspace{-2pt}
\section{Experiments}
\subsection{Datasets and Implementation Details}
\noindent \textbf{Modalities and datasets.} We evaluate UniBind on seven modalities -- image, point cloud, audio, thermal, video, event, and text. 
For each modality, we assess our UniBind on two mainstream datasets at least. 
A summary of the datasets utilized is presented in Table~\ref{Table_dataset_1}.


\noindent \textbf{Multi-modal backbone models.} Since our UniBind can be flexibly applied to the existing CLIP-style multi-modal learning models, in this paper, we implement UniBind with ImageBind~\cite{girdhar2023imagebind}, PointBind~\cite{guo2023point}, CLIP~\cite{radford2021learning}, E-CLIP~\cite{zhou2023clip}, Audio-CLIP~\cite{guzhov2022audioclip}, and Point-CLIP~\cite{zhang2022pointclip}. 
The backbone models are kept frozen and the linear layers at the end of visual encoders are updated during the LLM-augmented contrastive learning. 

\noindent \textbf{Knowledge Base.} We generate 1,000 descriptions for each category name via LLMs (GPT-4~\cite{openai2023gpt4}, LLaMa~\cite{touvron2023llama}) and generate multi-modal data descriptions via multi-modal LLMs (BLIP-2~\cite{li2023blip}, LLaMa-Adapter~\cite{zhang2023llama}).
We construct our knowledge base with these two sets of description texts.

\begin{table}[t!]
\renewcommand{\tabcolsep}{4pt}
\resizebox{\linewidth}{!}{
\begin{tabular}{lcccc}
\toprule
\multicolumn{1}{c|}{Model} & \multicolumn{2}{c|}{\textcolor{blue!70!black}{Image}} & \multicolumn{2}{c}{\textcolor{red!70!black}{Event}/\textcolor{magenta!70!black}{Audio}/\textcolor{yellow!70!black}{PC}} \\ \cmidrule{2-5}
&\multicolumn{1}{|c|}{\textcolor{blue!70!black}{IN-1K}} & \multicolumn{1}{c|}{\textcolor{blue!70!black}{Place-365}} & \multicolumn{1}{c|}{Dataset 1} & \multicolumn{1}{c}{Dataset 2} \\ \midrule
CLIP~\cite{radford2021learning} & 68.30 & 29.95 & \tiny{\XSolidBrush} & \tiny{\XSolidBrush} \\
\textbf{Ours w/ CLIP} & 78.63 & 39.14 & \tiny{\XSolidBrush} & \tiny{\XSolidBrush} \\
$\Delta $ & {\textbf{+10.35}} & {\textbf{+9.19}} & - & - \\ \midrule
E-CLIP~\cite{zhou2023clip} & 68.30 & 29.95 & 50.40 & 4.13 \\
\textbf{Ours w/ E-CLIP} & 78.63 & 39.14&  54.26 & 7.91 \\
$\Delta $ & {\textbf{+10.35}} & {\textbf{+9.19}} & {\textbf{+3.83}} & {\textbf{+3.78}} \\ \midrule
Audio-CLIP~\cite{guzhov2022audioclip} & 40.51 & 18.76 & 68.60 & 68.78 \\
\textbf{Ours w/ Audio-CLIP} & 46.44 & 22.60 & 71.25 & 69.52 \\
$\Delta $ & {\textbf{+5.93}} & {\textbf{+3.84}} & {\textbf{+2.65}} & {\textbf{+0.74}} \\ \midrule
Point-CLIP~\cite{zhang2022pointclip} & 59.60 & 25.56 & 20.20 & 89.20 \\
\textbf{Ours w Point-CLIP}~\cite{zhang2022pointclip} & 62.58 & 27.10 & 21.43 & 90.27 \\
$\Delta $ & {\textbf{+2.98}} & {\textbf{+1.54}} & {\textbf{+1.23}}  & {\textbf{+1.07}} \\
\bottomrule
\end{tabular}}
\caption{Emergent zero-shot recognition in image + X modalities. Dataset 1 and 2 indicate \textit{N-Cal} and \textit{N-1N-1K}, \textit{ESC-50} and \textit{Urban-S}, and \textit{ModelNet40} and \textit{ShapeNet}, respectively.} 
\vspace{-10pt}
\label{Table_main_2}
\end{table}
\subsection{Zero-shot Recognition}

\noindent \textbf{Settings.} The emergent zero-shot recognition is first proposed in ImageBind~\cite{girdhar2023imagebind} which means just by pre-training on (image, text) and (image, audio).
As shown in Tab.~\ref{Table_dataset_1}, we evaluate UniBind in \textbf{12} main-stream datasets from \textbf{6} modalities. We directly test the recognition performance without training (\textit{more details can be found in the suppl.}). 

\noindent \textbf{Results.} 
We evaluate the zero-shot recognition performance in comparison to existing methods across various modalities, including image, point cloud, audio, thermal, video, and event. In Tab.~\ref{Table_main_1}, we present the performance results of our approach when applied with CLIP-style multi-modal learning models that align more than three modalities. Additionally, we show the performance of our UniBind in conjunction with two-modality methods in Tab.~\ref{Table_main_2}.
Our UniBind significantly improves the performances of CLIP-style multi-modal models.
Across all benchmarks, UniBind achieves large gains about an average of +6.27\% in top 1 accuracy and even compares favorably to supervised specialist models trained for the special modality and task.

\subsection{Fine-tuning Recognition}
\noindent \textbf{Settings.} We follow ImageBind~\cite{girdhar2023imagebind} and only train the linear layer after the frozen visual encoders with the training dataset, and then evaluate our UniBind on the testing dataset with the same metric in the zero-shot setting.

\noindent \textbf{Results.} In Tab.~\ref{Table_main_1}, we compare our approach with the supervised methods that use {\fontfamily{qcr}\selectfont ["A photo of a [category name]"]} as the text label during the training stage. Our UniBind outperforms the supervised method on 10 benchmarks spanning 6 modalities, exhibiting an average improvement of +1.26\%. In particular, our UniBind shows more significant gains in the datasets containing a large number of categories, such as +6.75\% in ImageNet (1,000 categories) and +7.54\% in Place-365 (365 categories). It demonstrates the advantages of our approach to apply in complex semantic data.

\vspace{-2pt}
\section{Ablation Study and Analysis}

\begin{figure}[t!]
    \centering
    \includegraphics[width=\linewidth]{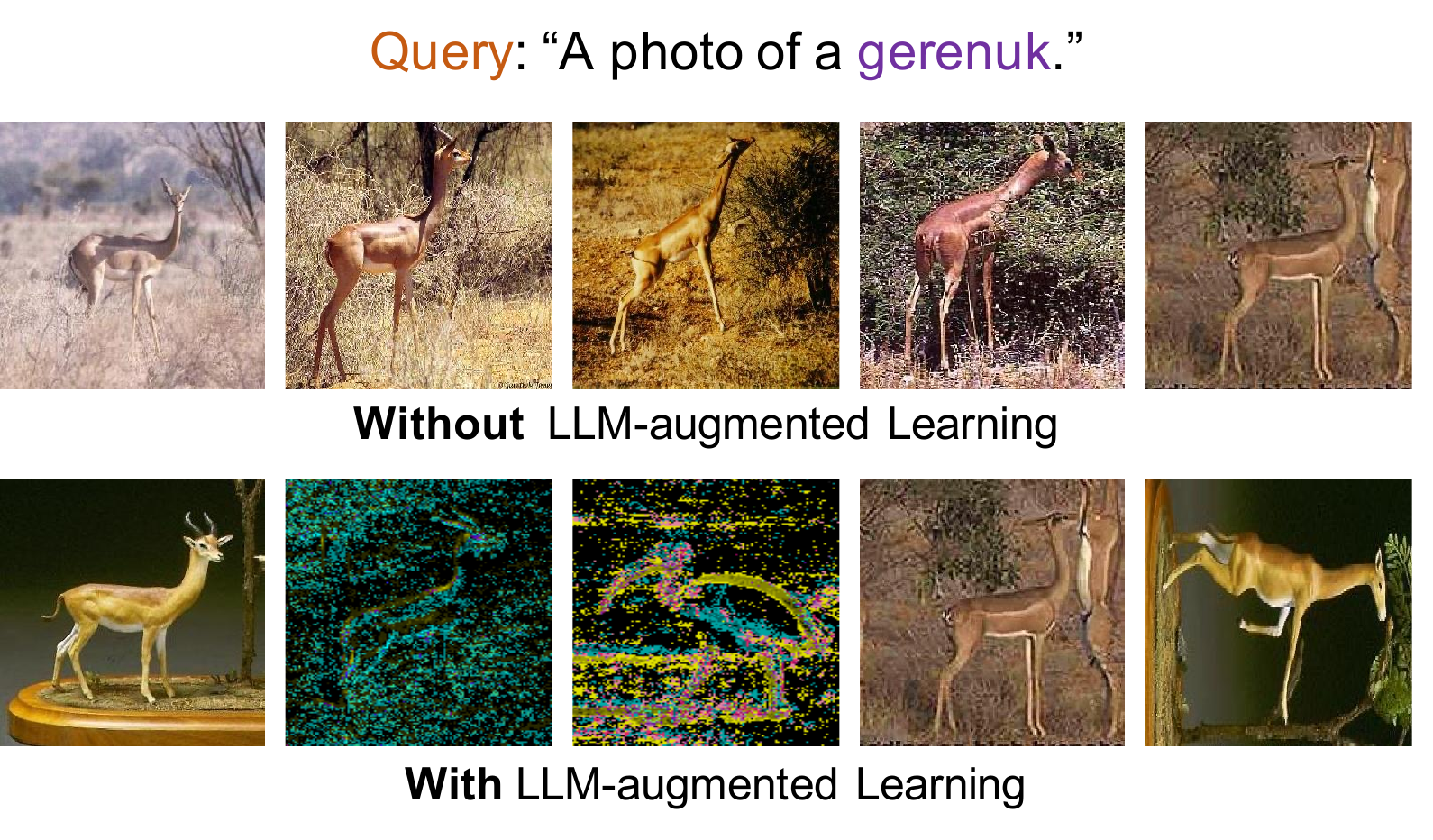}
    \vspace{-22pt}
    \caption{Top 5 results from text to events \& images retrieval. We choose {\fontfamily{qcr}\selectfont ["A photo of a [category]"]} as the query to retrieve events and images in the same embedding space.}
    \vspace{-8pt}
    \label{retrieval}
\end{figure}

\subsection{LLM-augmented Contrastive Learning}
To investigate the effectiveness of our proposed LLM-augmented contrastive learning, we conduct ablation studies on the cross-modal retrieval task. We experiment with E-CLIP~\cite{zhou2023clip} and PointBind~\cite{guo2023point} and subsequently report the results for event-to-image retrieval and image-to-event retrieval in Tab.~\ref{Table_ab_LG_2}. 
The recall score improvement increases incrementally from the top 1 to the top 20, illustrating the efficacy of our approach in aligning modalities with the same semantics. In addition, as shown in Fig.~\ref{retrieval}, we demonstrate the case of cross-modal retrieval based on PointBind~\cite{guo2023point} adapted event modality. In this case, we use the text {\fontfamily{qcr}\selectfont["A photo of a Gerenuk."]} to retrieve images and events in the same representation space. In the absence of LLM-augmented contrastive learning, the top 5 retrieval results solely consist of images. By contrast, with LLM-augmented contrastive learning, the retrieval results are more balanced across image and event modalities.

\begin{table}[t!]
\renewcommand{\tabcolsep}{1pt}
\resizebox{\linewidth}{!}{
\begin{tabular}{lcccccc}
\toprule
\multicolumn{1}{c|}{Model} & \multicolumn{3}{c|}{\textcolor{blue!70!black}{Image}-to-\textcolor{red!70!black}{Event}}& \multicolumn{3}{c}{\textcolor{red!70!black}{Event}-to-\textcolor{blue!70!black}{Image}}  \\ \cmidrule{2-7}
\multicolumn{1}{l|}{} & \multicolumn{1}{c|}{R@1} & \multicolumn{1}{c|}{R@10} & \multicolumn{1}{c|}{R@20} & \multicolumn{1}{c|}{R@1} & \multicolumn{1}{c|}{R@10} & \multicolumn{1}{c}{R@20}  \\ \midrule
E-CLIP~\cite{zhou2023clip} & 79.52 & 93.08 & 95.51 & 76.29 & 91.80 & 94.61 \\
\textbf{E-CLIP w LCL}& 78.95 & 94.32 & 97.06 & 77.04 & 93.62 & 96.70 \\
$\Delta $ & -0.57 & \textbf{+1.24} & \textbf{+1.55} & \textbf{+0.75} & \textbf{+1.82} & \textbf{+2.09}  \\ \midrule
PointBind (\textbf{+Event})~\cite{guo2023point} & 14.07 & 40.79 & 49.46 & 9.00 & 29.32 & 37.70  \\
\textbf{PointBind w LCL} & 14.12 & 41.25 & 50.98 & 14.29 & 44.34 & 55.66\\
$\Delta $ & \textbf{+0.05} & \textbf{+0.46} & \textbf{+1.52} & \textbf{+5.29} & \textbf{+15.02} & \textbf{+17.96}  \\
\bottomrule
\end{tabular}}
\caption{Multi-modal retrieval result with/without LLM-augmented Contrastive Learning (LCL). We evaluate E-CLIP~\cite{zhou2023clip} and PointBind~\cite{guo2023point} in Image-to-Event and Event-to-Image tasks.}
\vspace{-6pt}
\label{Table_ab_LG_2}
\end{table}

\begin{figure}[t!]
    \centering
    \includegraphics[width=0.48\textwidth]{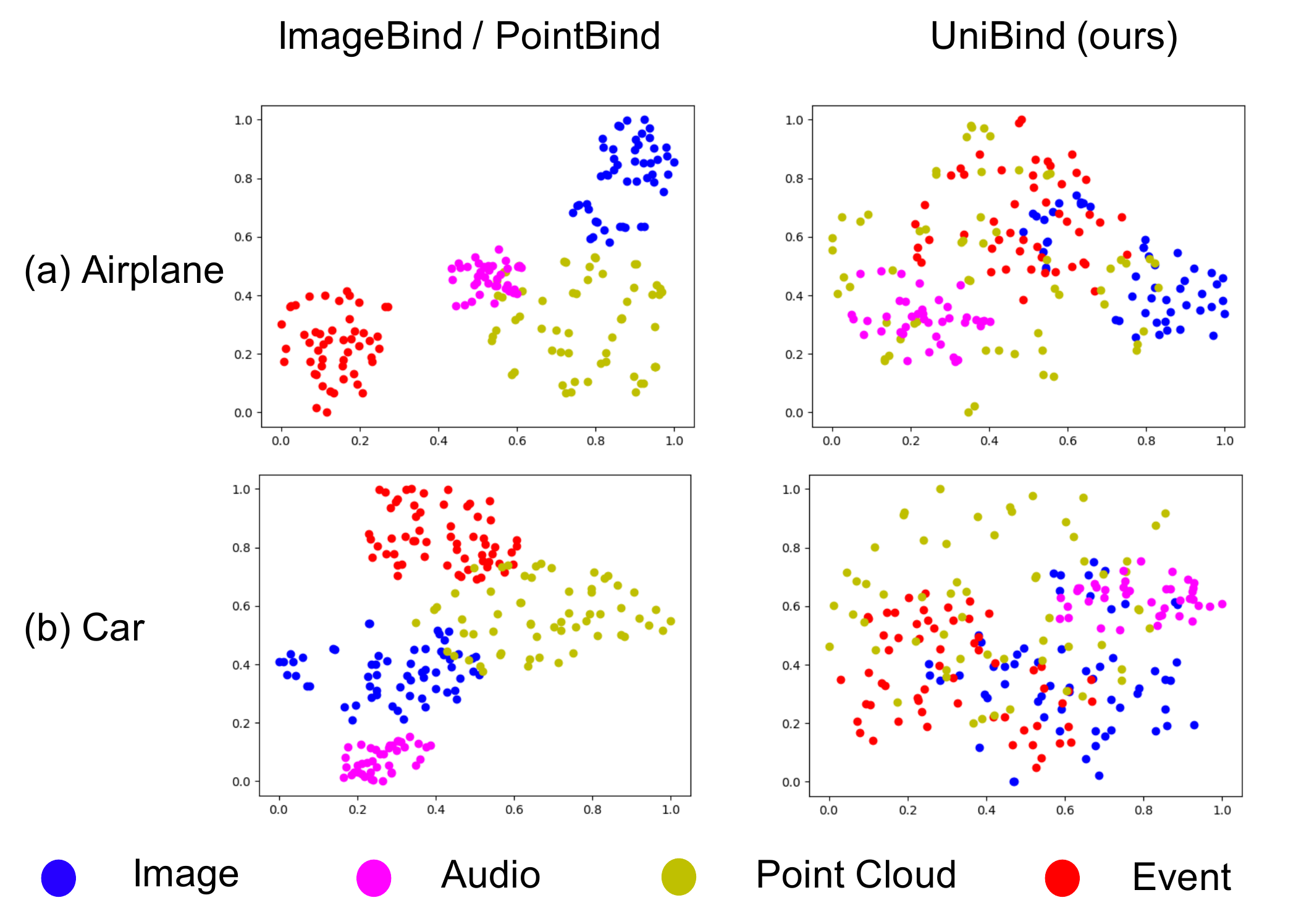}
    \caption{Representation space visualization of ImageBind / PointBind and our UniBind. We sample 64 data in the same semantic for each modality, specifically, {\fontfamily{qcr}\selectfont [‘Airplane']} data representation space shown in (a), {\fontfamily{qcr}\selectfont [‘Car']} data shown in (b).}
    \vspace{-8pt}
    \label{rep_space}
\end{figure}

Furthermore, the results of the t-SNE visualization in Fig.~\ref{rep_space} reveal the differences between the representation spaces constructed by ImageBind~\cite{girdhar2023imagebind} / PointBind~\cite{guo2023point} and our UniBind. For example, in Fig.~\ref{rep_space} (a), We select embeddings with the same semantic label {\fontfamily{qcr}\selectfont [‘airplane']} from image, audio, point cloud, and event modalities, and visualize 64 randomly chosen samples from each modality. In the representation space, embeddings from different modalities tend to cluster around their respective modalities. Thereby, with LLM-augmented contrastive learning, multi-modal embeddings cluster around the same semantic label in our unified modality-agnostic representation space.

\subsection{Embedding Center Localization}

\begin{figure}[t!]
    \centering
    \includegraphics[width=0.98\linewidth]{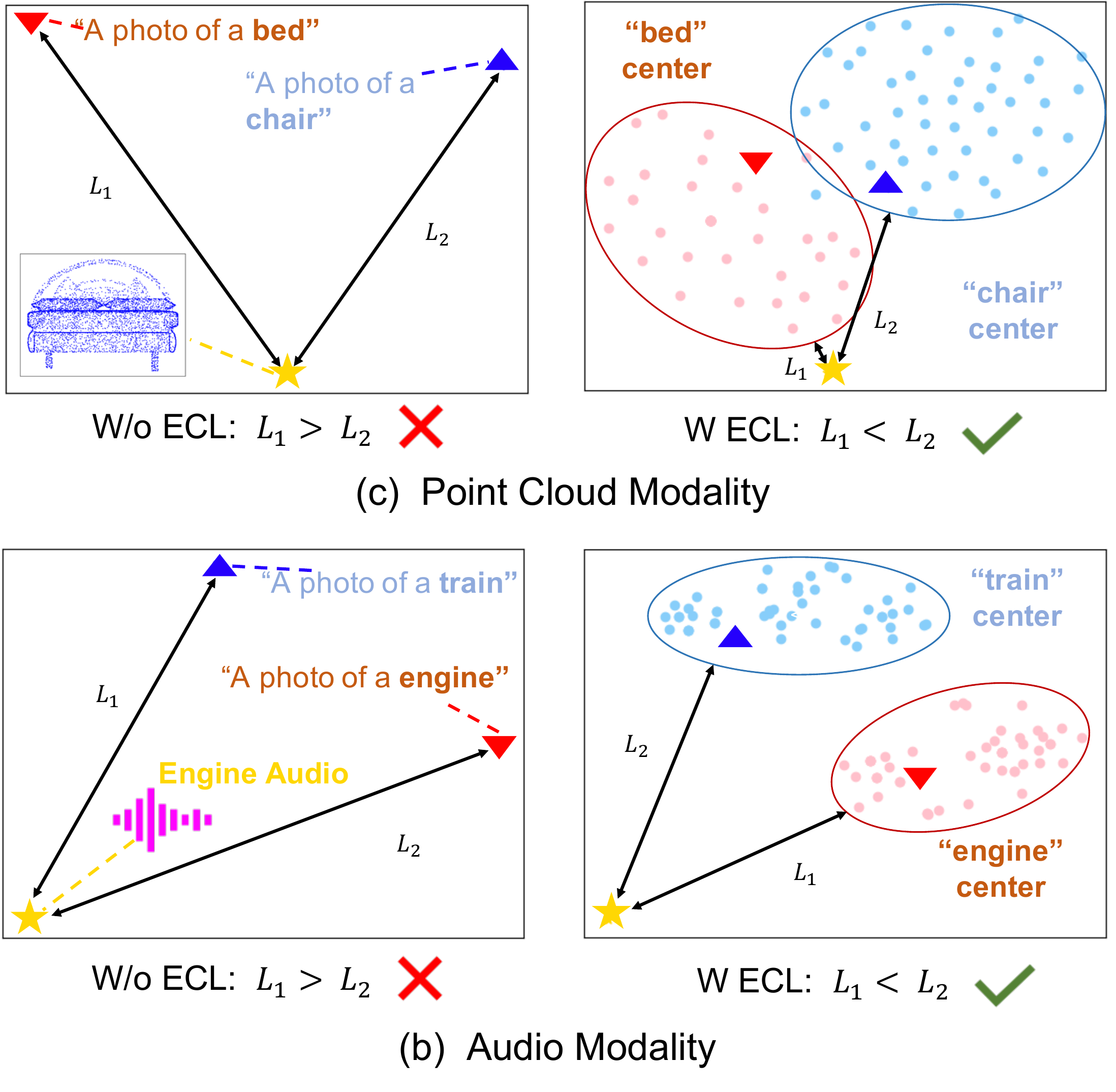}
    \vspace{-12pt}
    \caption{Embedding centers t-SNE visualization in the point cloud and audio modalities. (a) in this case, we simply use the prompts as the centers of both the bed and chair categories; on the right, we use our localized embedding centers. (b) show the cases in audio modality.}
    \label{ecl_ab}
\end{figure}

\begin{table}[t!]
\renewcommand{\tabcolsep}{3pt}
\resizebox{\linewidth}{!}{
\begin{tabular}{lcccc}
\toprule
Modality & \textcolor{blue!70!black}{Image} & \textcolor{magenta!70!black}{Audio} & \textcolor{yellow!70!black}{PointCloud}& \textcolor{red!70!black}{Event}\\  \cmidrule{1-5}
E-CLIP~\cite{zhou2023clip}  & 68.30 &\tiny{\XSolidBrush} & \tiny{\XSolidBrush}& 50.40 \\ 
\textbf{E-CLIP w ECL} & 78.63 &\tiny{\XSolidBrush} & \tiny{\XSolidBrush}& 54.26 \\ 
$\Delta $  &\textbf{+10.33} & -& -& \textbf{+3.86} \\ \midrule
AudioCLIP ~\cite{guzhov2022audioclip} & 40.51 &68.60 & \tiny{\XSolidBrush}& \tiny{\XSolidBrush}\\ 
\textbf{AudioCLIP w ECL} & 46.44 &71.25  & \tiny{\XSolidBrush}&\tiny{\XSolidBrush}\\ 
$\Delta $  & \textbf{+5.93}& \textbf{+2.65}& & \\ \midrule
ImageBind ~\cite{girdhar2023imagebind} & 77.70 &66.90  &\tiny{\XSolidBrush} &\tiny{\XSolidBrush} \\ 
\textbf{ImageBind w ECL} &83.25 &71.70 & \tiny{\XSolidBrush}&\tiny{\XSolidBrush}  \\ 
$\Delta $  &\textbf{+5.55} & \textbf{+4.80} & -& -\\ \midrule
PointBind (+Event)~\cite{guo2023point} &77.70 &66.90 &77.67 & 50.98\\ 
\textbf{PointBind w ECL} & 83.25 & 71.70 & 80.59 & 59.26    \\
$\Delta $  &\textbf{+5.55}  & \textbf{+4.80} & \textbf{+2.92} & \textbf{+8.28}\\ 
\bottomrule
\end{tabular}}
\caption{Performance in zero-shot recognition task with/without Embedding Center Localization (ECL) with five multi-modal models in four modalities.}
\vspace{-12pt}
\label{Table_ab_ECR_1}
\end{table}

\begin{figure}[t!]
    \centering
    \includegraphics[width=\linewidth]{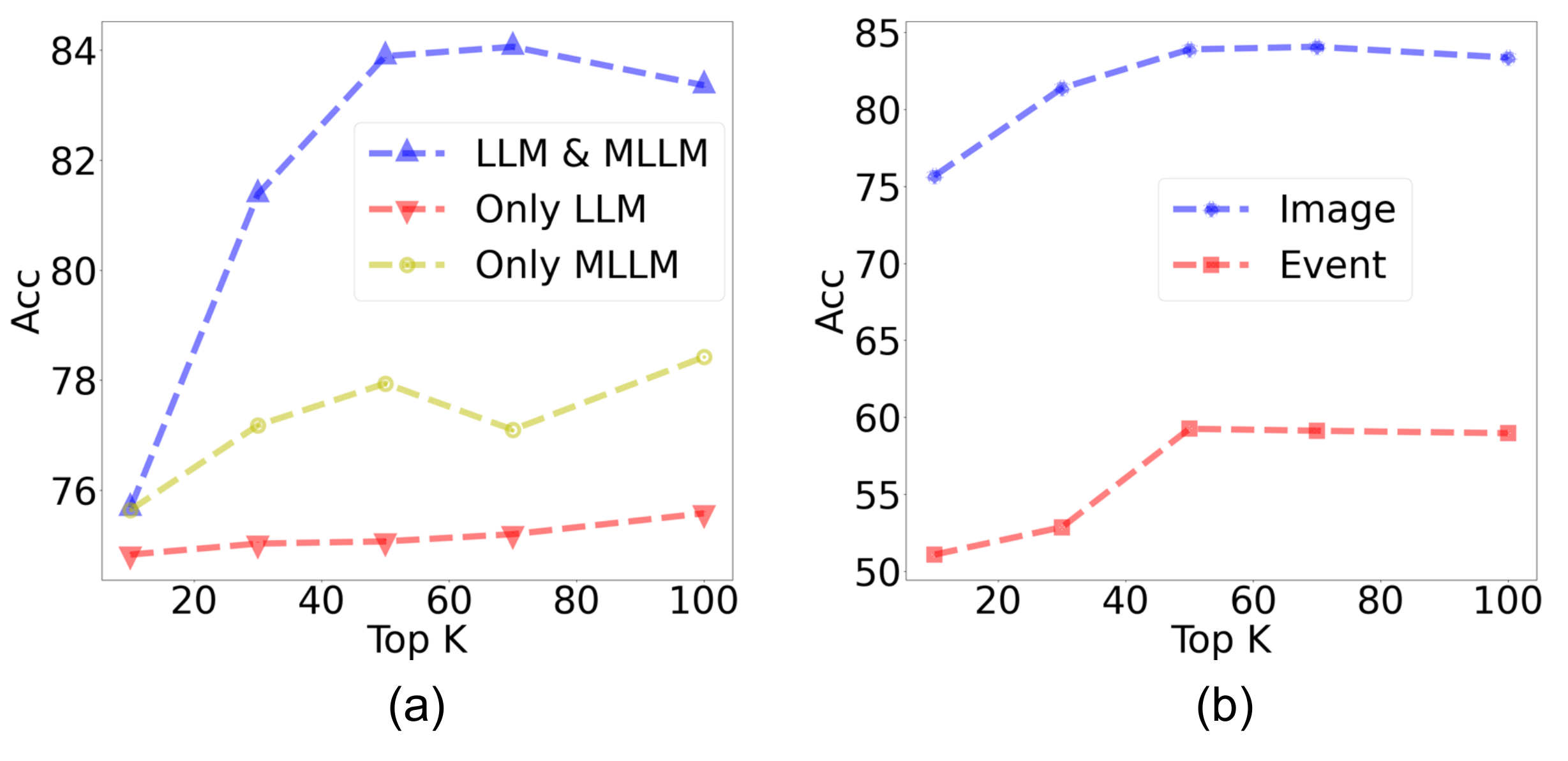}
    \vspace{-10pt}
    \caption{Ablation study of the knowledge base. (a) compares three ways to build a knowledge base by only LLMs, only Multi-modal LLMs, and both LLMs and Multi-modal LLMs. (b) shows the performance of selecting the top 10-100 texts for each category in image and event modalities.}
    \label{ab_of_topk}
\end{figure}

The results presented in Tab.~\ref{Table_ab_ECR_1} demonstrate the effect of improving the zero-shot recognition performance with E-CLIP~\cite{zhou2023clip}, AudioCLIP~\cite{guzhov2022audioclip}, ImageBind~\cite{girdhar2023imagebind}, and PointBind~\cite{guo2023point} on four modalities. Our embedding center localization method attains a substantial improvement, averaging +6.50\% when applied to these CLIP-style multi-modal learning models. As the t-SNE visualization in Fig.~\ref{ecl_ab} illustrated, our approach results in more distinct semantic boundaries between different categories, effectively enhancing recognition accuracy and reducing interference from other categories.
We also study the impact of LLMs and Multi-modal LLMs, while further examining the optimal number of selected texts for each category. We compare our UniBind to the method that only selects texts from LLMs or only from Multi-modal LLMs to construct the knowledge base and present the results in Fig.~\ref{ab_of_topk}. Evidently, the knowledge base created by LLMs and Multi-modal LLMs exhibits the best performance, with the selection of the top 50 texts for each category proving to be the most effective choice for localizing the embedding centers.
Lastly, we present the results of comparisons with other language-augmented methods in Tab.~\ref{Table_ab_ECR_2}. Our UniBind demonstrates the best performance.

\begin{table}[t!]
\renewcommand{\tabcolsep}{3pt}
\resizebox{\linewidth}{!}{
\begin{tabular}{lcccc}
\toprule
\multirow{1}{*}{Method} & \multicolumn{1}{c}{\textcolor{blue!70!black}{Image}} & \multicolumn{1}{c}{\textcolor{yellow!70!black}{PointCloud}} & \multicolumn{1}{c}{\textcolor{violet}{Video}} & \multicolumn{1}{c}{\textcolor{red!70!black}{Event}} \\ \cmidrule{1-5}
Simple Prompts~\cite{radford2021learning} & 75.86 & 76.02 & 30.92 & 50.39\\
Complex Prompts~\cite{radford2021learning, guo2023point}  & 77.70 & 77.67 & 31.27 & 50.98 \\
Word-net Augmented ~\cite{hu2021knowledgeable} & 78.10 & 79.09 & \textbf{41.03} & 50.60 \\
LLM-generated Prompts~\cite{zhu2022pointclip} & 79.59 & 77.43 & 34.13 & 51.93 \\
Ours & \textbf{83.25} & \textbf{80.59} & 40.90 & \textbf{59.26}\\
\bottomrule
\end{tabular}}
\caption{Performance of various language-augmented methods in the zero-shot recognition task. We compare our LLM-augmented method with simple promotes by default of CLIP~\cite{radford2021learning}, complex prompts used by PointBind~\cite{guo2023point}, word-net augmented prompts~\cite{hu2021knowledgeable}, and LLM-generated prompts.}
\label{Table_ab_ECR_2}
\vspace{-6pt}
\end{table}
\section{Conclusion}
In this paper, we proposed UniBind, a multi-modal learning approach that renders the alignment centers modality-agnostic and further learns a unified and balanced representation space, empowered by the large language models (LLMs) and the multi-modal large language models (multi-modal LLMs). Our UniBind achieves remarkable performance boosts and is compatible with all CLIP-style multi-modal learning models. Additionally, we examined the potential of LLMs and Multi-modal LLMs for multi-modal representation space learning. 

\noindent \textbf{Limitations and Future Works.}
The robustness of the LLM-augmented method requires enhancement. In response, our future work will concentrate on harnessing the capabilities of LLMs to augment the robustness of the modality-agnostic representation space.

\noindent\textbf{Acknowledgement.}
This paper is supported by the National Natural Science Foundation of China (NSF) under Grant No. NSFC22FYT45 and the Guangzhou City, University and Enterprise Joint Fund under Grant No.SL2022A03J01278.

\clearpage
{
    \small
    \bibliographystyle{ieeenat_fullname}
    \bibliography{main}
}

\newpage
\appendix

\section{Construction of Knowledge Base}
\label{Construction of Knowledge Base}
The knowledge base consists of two components: \textbf{1)} Category descriptions, generated by large language models (LLMs). \textbf{2)} Multi-modal data descriptions produced by multi-modal large language models (multi-modal LLMs).

Illustrating with ImageNet-1k~\cite{deng2009imagenet} as an example, we initially generate descriptive texts for 1,000 categories using GPT-4~\cite{openai2023gpt4} and LLaMA~\cite{touvron2023llama}. For each category, we generate 1,000 descriptive texts, limiting the output sequence's maximum length to 77 tokens. Specific instances are detailed in Sec.\ref{LLM}. Subsequently, we generate descriptions for the visual data. In the case of ImageNet-1k\cite{deng2009imagenet}, we generate descriptions for each image in the dataset using BLIP-2~\cite{li2023blip}, ensuring the sequence length remains below 77 tokens. Concrete examples are provided in Sec.\ref{MLLM}. Finally, the organizational structure of our knowledge base is delineated in Sec.\ref{organiza_structure}.

\subsection{Cases of Generation via LLMs}
\label{LLM}
\begin{figure}[t!]
    \centering
    \includegraphics[width=\linewidth]{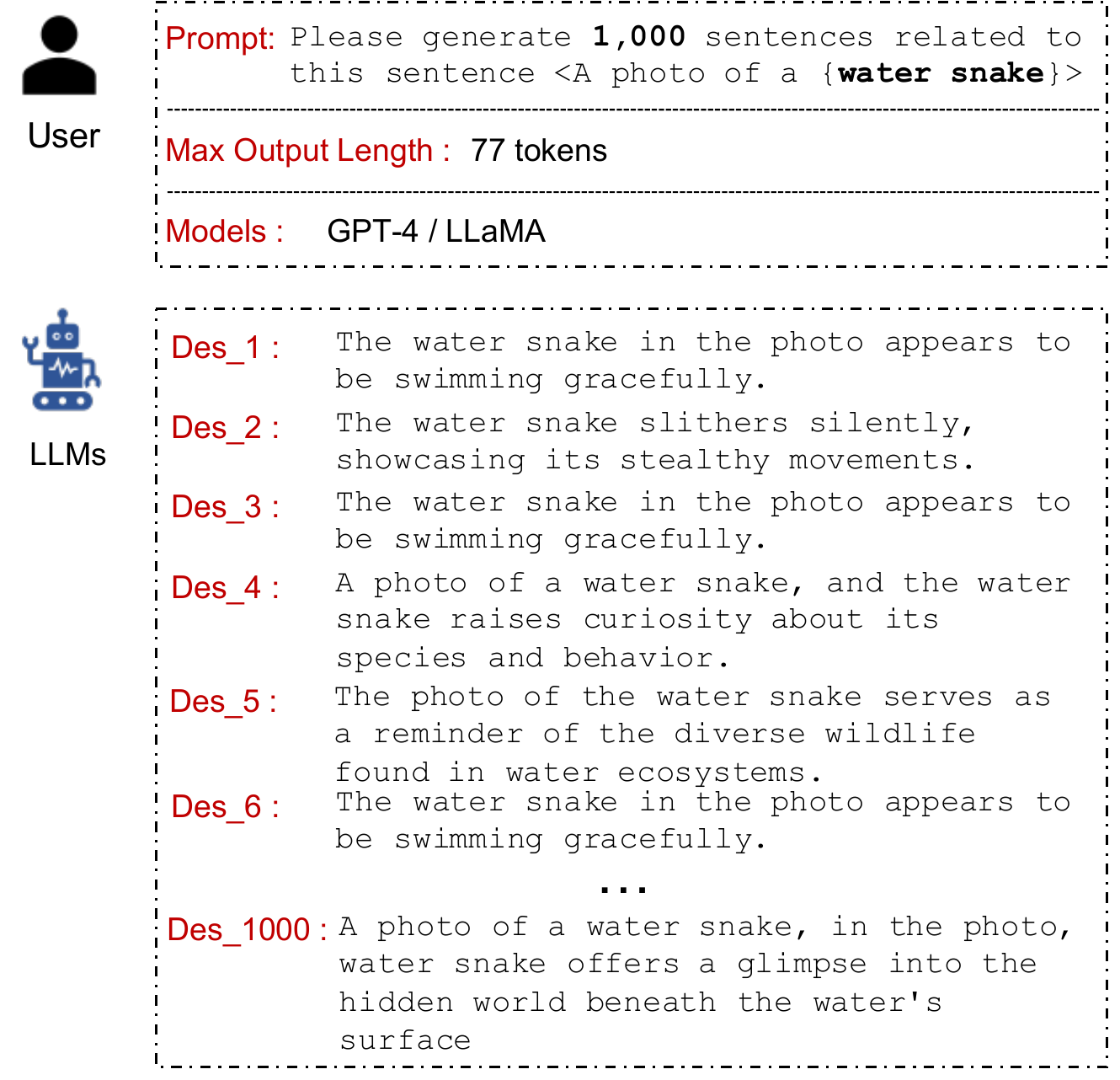}
    \caption{The case of description generation via LLMs. We show the generated descriptions for the category {\fontfamily{qcr}\selectfont [water snake]}.}
    \label{case_1}
\end{figure}

We employ GPT-4~\cite{openai2023gpt4} and LLaMA~\cite{touvron2023llama} to generate category descriptions. Illustrated in Fig~\ref{case_1}, we generate 1,000 descriptions for the category {\fontfamily{qcr}\selectfont [water snake]}. To achieve this, we utilize the prompt {\fontfamily{qcr}\selectfont ["Please generate 1,000 sentences related to this sentence <A photo of a {water snake}>"]} as input, facilitating the generation of effective descriptions for the localization of the embedding center.

\subsection{Cases of Generation via Multi-modal LLMs}
\label{MLLM}
\begin{figure}[t!]
    \centering
    \includegraphics[width=\linewidth]{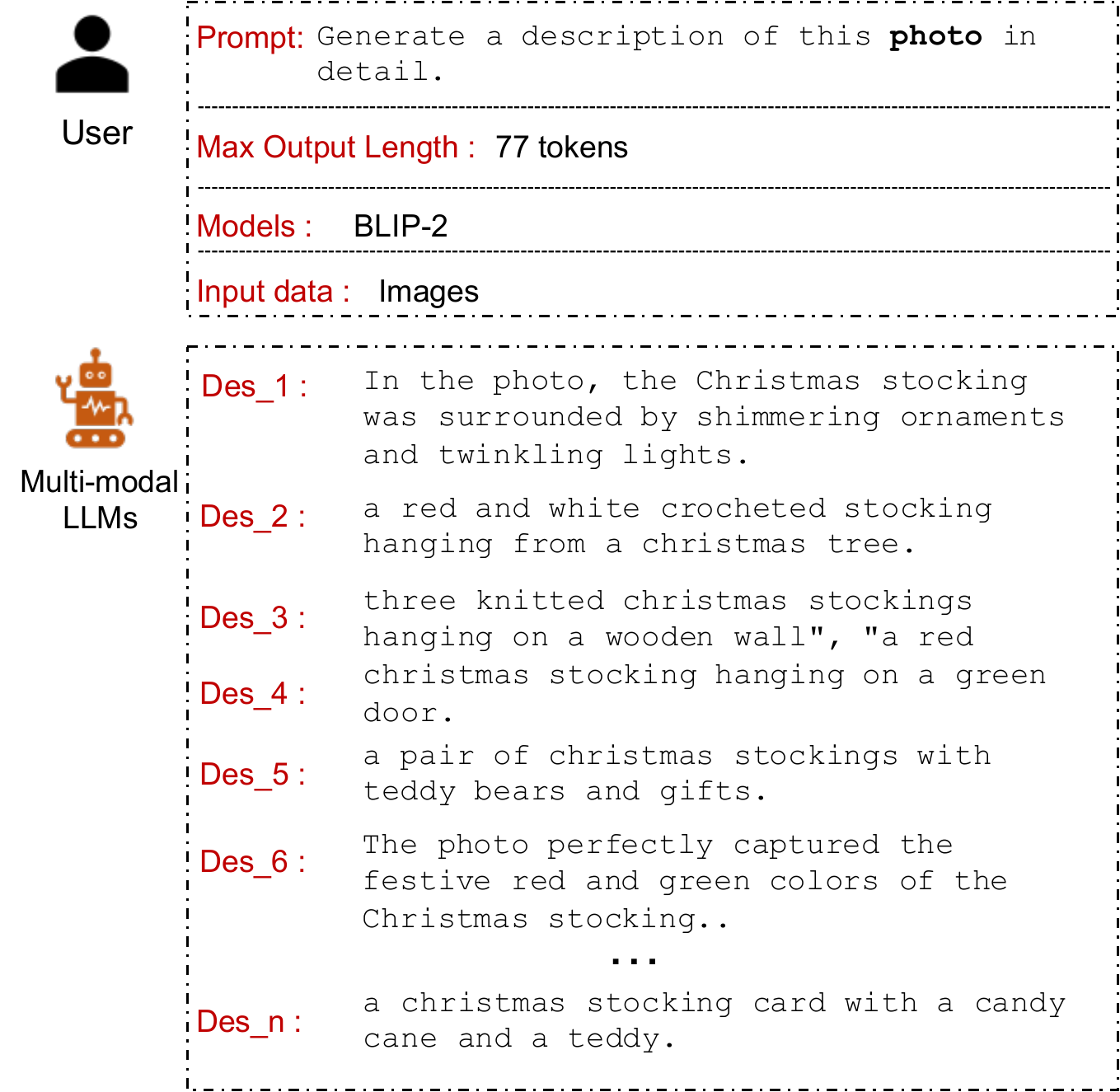}
    \caption{The case of generation via BLIP-2~\cite{li2023blip}. We present the generated descriptions for ImageNet-1k~\cite{deng2009imagenet}.}
    \label{case_2}
\end{figure}

For the image, event, and thermal modalities, we produce multi-modal data descriptions using BLIP-2~\cite{li2023blip}. Specifically, we employ paired RGB images from event and thermal data to generate these descriptions. The process of generating descriptions for image data is illustrated in Fig.~\ref{case_2}. We utilize the prompt {\fontfamily{qcr}\selectfont [Generate a detailed description of this \textbf{photo}]} as the text input, while the visual inputs consist of RGB images.

\begin{figure}[t!]
    \centering
    \includegraphics[width=\linewidth]{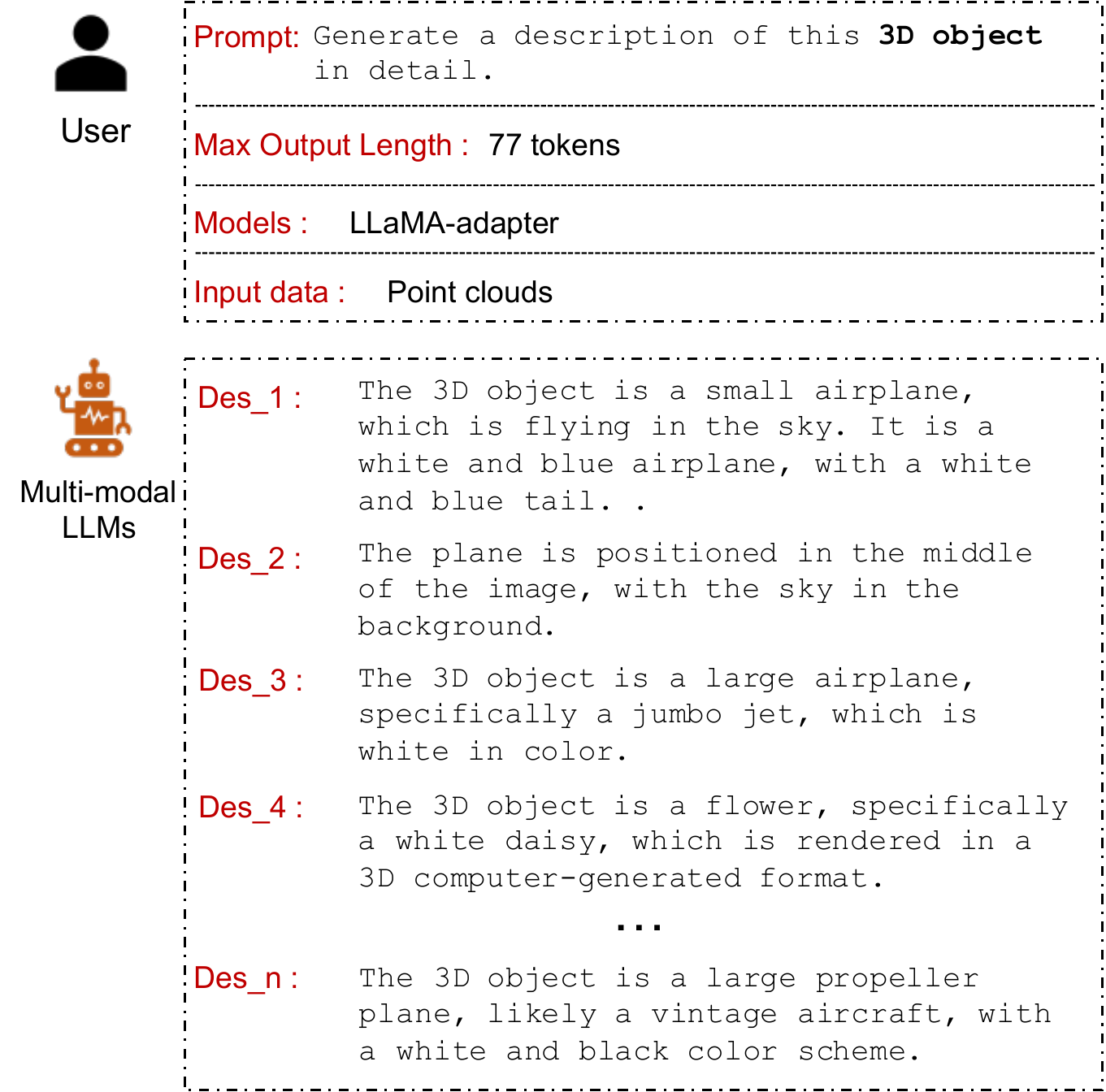}
    \caption{The case of generation via LLaMA-adapter~\cite{zhang2023llama}. We show the generation descriptions for ModelNet-40~\cite{wu20153d}}
    \label{case_3}
\end{figure}

For the audio, video, and point cloud modalities, we employ the LLaMA-adapter~\cite{zhang2023llama} to generate descriptions for multi-modal data. Illustrated in Fig.~\ref{case_3}, we present a case of generating descriptions for point cloud data. In this instance, we utilize the prompt {\fontfamily{qcr}\selectfont [Generate a detailed description of this \textbf{3D object}]} as the text input, with the visual inputs consisting of point cloud data.

\subsection{Organizational Structure}
\label{organiza_structure}
\begin{figure}[t!]
    \centering
    \includegraphics[width=\linewidth]{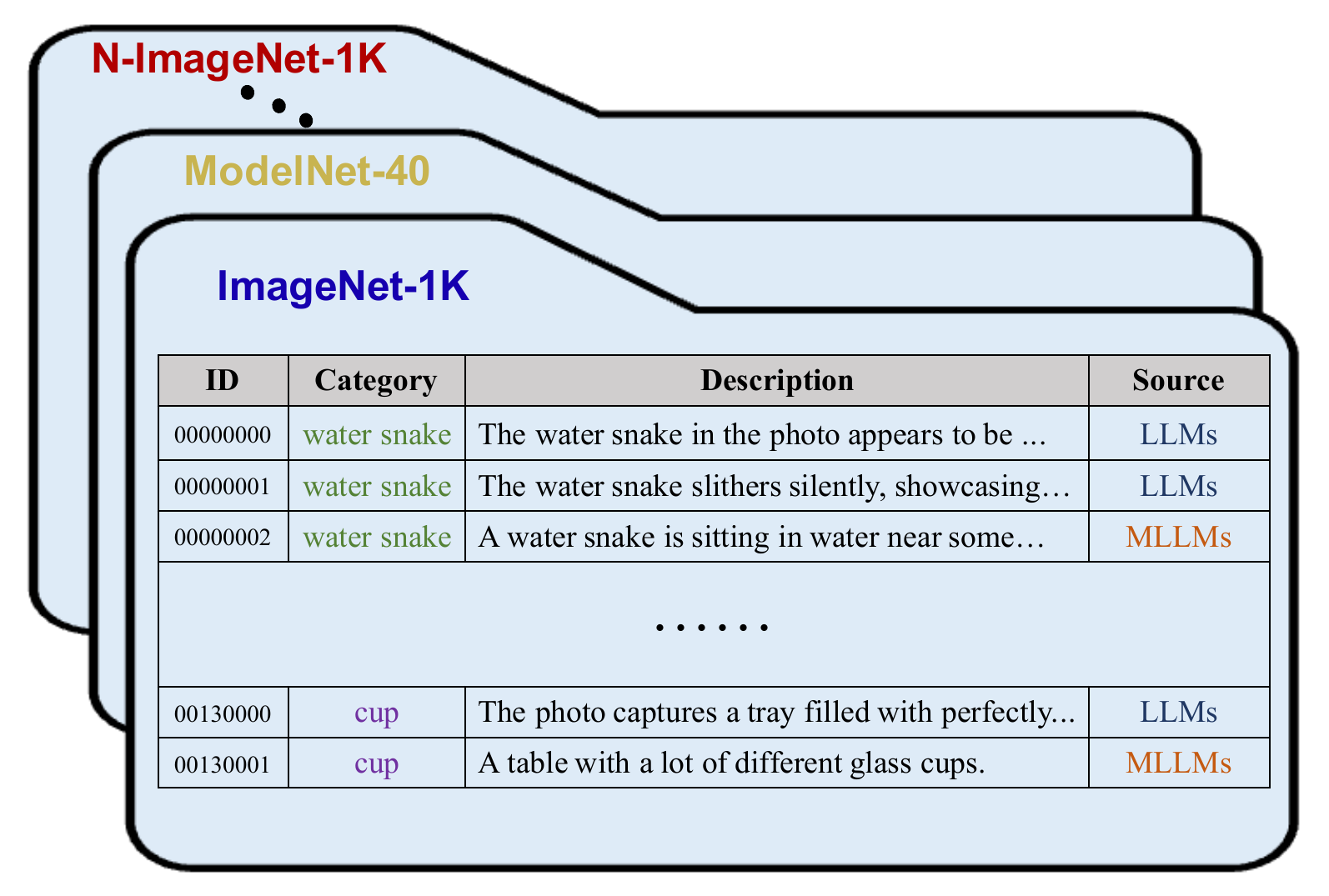}
    \caption{The organizational structure of our knowledge base.}
    \label{structure}
\end{figure}

Lastly, we present the organizational structure of our knowledge base in Fig.~\ref{structure}. The knowledge base arranges descriptions generated from the same dataset in a table, with each table featuring four keys: \textit{ID}, \textit{Category}, \textit{Description}, and \textit{Source}. Descriptions with the same \textit{Category} key value are selected to localize embedding centers for categories. During training, we retrieve paired descriptions of input visual data using the \textit{ID} key.

\section{Implementation Details}
\label{Implementation Details}

\subsection{Datasets and Metrics}
We experiment with our method on 13 datasets. Next, we show the details and metrics of these benchmark datasets.

\noindent \textcolor{blue!70!black}{\textbf{ImageNet-1K (IN-1K)}}~\cite{deng2009imagenet}. 
It is a standard image dataset designed for recognition tasks encompassing 1,000 categories. It serves the dual purpose of both training and evaluation. In the zero-shot setting, we assess both baseline models and our proposed method on the test set without training. Accuracy is employed as the metric for evaluation.

\noindent \textcolor{blue!70!black}{\textbf{Places-Stanford-365 (P365)}} ~\cite{lopez2020semantic}.
The Stanford-365 dataset is designed for scene recognition, comprising 365 categories. The evaluation setting for this dataset mirrors that of ImageNet-1K~\cite{deng2009imagenet}.

\noindent \textcolor{blue!70!black}{\textbf{Caltech-101 (cal)}} ~\cite{fei2004learning}.
The Caltech-101 dataset is a well-established benchmark dataset within the domain of computer vision, tailored specifically for object recognition tasks. Comprising images from 101 distinct object categories, it captures diverse scenes from real-world settings. In this study, we employ the dataset for evaluating models in the context of object recognition. Additionally, we utilize both the Caltech-101 and N-Caltech-101 datasets for cross-modal retrieval tasks. Accuracy is employed as the metric for the recognition task, while recall serves as the metric for the cross-modal retrieval task.

\noindent \textcolor{yellow!70!black} {\textbf{ModelNet-40 (ModelNet40)}}~\cite{wu20153d}.
ModelNet-40 serves as a widely adopted benchmark dataset within the realm of 3D object recognition. Encompassing 40 object categories, it includes items such as chairs, tables, airplanes, cars, and various household objects. Each category is adequately represented with a substantial number of instances, ensuring a comprehensive and representative sample. In the evaluation of recognition, accuracy is employed as the metric.

\begin{table}[ht!]
\renewcommand{\tabcolsep}{4pt}
\resizebox{\linewidth}{!}{
\begin{tabular}{llccc}
\toprule
\multicolumn{1}{c}{Modalities} & \multicolumn{1}{c}{Dataset} & \multicolumn{1}{c}{batch size} & \multicolumn{1}{c}{lr} & \multicolumn{1}{c}{total epochs}\\ \midrule
\multirow{3}{*}{\textcolor{blue!70!black}{Image}} & \textcolor{blue!70!black}{ImageNet-1K (IN-1K)}~\cite{deng2009imagenet} & 1,024 & 5e-4 & 15\\ 
& \textcolor{blue!70!black}{Places-Stanford-365 (P365)} ~\cite{lopez2020semantic}  & 1,024 & 5e-5 &20 \\
& \textcolor{blue!70!black}{Caltech-101 (cal)} ~\cite{fei2004learning}  & 128 & 1e-4 & 20\\ \midrule
\multirow{2}{*}{\textcolor{yellow!70!black}{PointCloud}} &\textcolor{yellow!70!black} {ModelNet-40 (ModelNet40)}~\cite{wu20153d}  & 128 & 5e-5 & 10\\
& \textcolor{yellow!70!black}{ShapeNet-part (ShapeNet)}~\cite{chang2015shapenet}  & 128 &  5e-5 & 10 \\ \midrule
\multirow{2}{*}{\textcolor{magenta!70!black}{Audio}} & \textcolor{magenta!70!black}{ESC 5-folds (ESC)}~\cite{piczak2015esc}  &  64 & 1e-4 & 10 \\
& \textcolor{magenta!70!black}{Urban-Sound-8K (Urban-S)}~\cite{salamon2014dataset} & 64 & 1e-4 & 10 \\ \midrule
\multirow{2}{*}{\textcolor{brown}{Thermal}} & \textcolor{brown}{LLVIP (LLVIP)} ~\cite{jia2021llvip} & 64 & 1e-3 & 20\\
& \textcolor{brown}{RGB-T Selected (RGB-T)} ~\cite{hwang2015multispectral} & 64 & 5e-3 & 20 \\ \midrule
\multirow{2}{*}{\textcolor{violet}{Video}} & \textcolor{violet}{MSR-VTT (MSR-VTT)}~\cite{xu2016msr}  & 128 & 5e-4 & 20\\
& \textcolor{violet}{UFC-101 (UFC-101)}~\cite{soomro2012ucf101} & 128 & 5e-4 & 20\\ \midrule
\multirow{2}{*}{\textcolor{red!70!black}{Event}} & \textcolor{red!70!black}{N-Caltech-101 (N-cal)}~\cite{orchard2015converting} & 128  & 1e-4 & 20\\
& \textcolor{red!70!black}{N-ImageNet-1K (N-IN-1K)}~\cite{kim2021n}& 1,024 & 5e-4 & 15 \\
\bottomrule
\end{tabular}}
\vspace{-5pt}
\caption{The hyperparameters of experiments with PointBind~\cite{guo2023point}. }
\vspace{-15pt}
\label{hypara}
\end{table}

\noindent \textcolor{yellow!70!black}{\textbf{ShapeNet-part (ShapeNet)}}~\cite{chang2015shapenet}.
ShapeNet-part stands as a prominent benchmark dataset extensively employed for 3D segmentation tasks. Within the scope of this work, we delineate the evaluation task on ShapeNet-part as a recognition task. The dataset comprises 16 categories of 3D objects. The evaluation metric employed for this task is accuracy.

\noindent \textcolor{magenta!70!black}{\textbf{ESC 5-folds (ESC)}}~\cite{piczak2015esc}.
The dataset comprises 2,000 audio clips of 5 seconds each, classified into 50 distinct classes. In the zero-shot setting, we employ the entire audio dataset to assess both baseline models and our proposed method. Conversely, in the fine-tuning setting, models are trained exclusively on the training set and subsequently evaluated on the test set. The metric employed for evaluation in both settings is accuracy.

\noindent \textcolor{magenta!70!black}{\textbf{Urban-Sound-8K (Urban-S)}}~\cite{salamon2014dataset}.
The UrbanSound8K dataset is a widely used collection of audio data designed for research in the field of urban sound recognition. UrbanSound8K consists of 8,732 audio clips, each lasting 4 seconds. These clips are extracted from longer field recordings and are labeled with specific sound classes. The dataset is annotated with 10 sound classes and we evaluate models on the test set with accuracy metric.

\noindent \textcolor{brown}{\textbf{LLVIP (LLVIP)}} ~\cite{jia2021llvip}.
The LLVIP dataset consists of RGB image and Thermal image pairs. We follow ImageBind~\cite{girdhar2023imagebind} to process it for a binary classification task. We crop out pedestrian bounding boxes and random bounding boxes (same aspect ratio and size as a pedestrian) to create a balanced set of 15,809 total boxes (7,931 `person' boxes). The metric used is top 1 accuracy.

\noindent \textcolor{brown}{\textbf{RGB-T Selected (RGB-T)}} ~\cite{hwang2015multispectral}.
We follow the processing methodology employed for LLVIP~\cite{jia2021llvip} in handling the RGB-T dataset. For a binary classification task, we specifically select 10,000 total bounding boxes, out of which 5,131 are labeled as 'person.' The top-1 accuracy is designated as the evaluation metric.

\noindent \textcolor{violet}{\textbf{MSR-VTT (MSR-VTT})}~\cite{xu2016msr}.
MSR-VTT contains a diverse set of videos covering a wide range of topics and scenarios. The dataset consists of 10,000 video clips from 20 categories. In this work, we evaluate multi-modal methods on the recognition task with these 20 categories. The metric used is accuracy.

\noindent \textcolor{violet}{\textbf{UFC-101 (UFC-101)}}~\cite{soomro2012ucf101}.
The UFC-101 dataset is a prevalent benchmark in the domain of action recognition. It encompasses a total of 13,320 video clips, portraying 101 distinct human action categories. For evaluation, accuracy is employed as the metric.

\noindent \textcolor{red!70!black}{\textbf{N-Caltech-101 (N-cal)}}~\cite{orchard2015converting}.
N-Caltech-101 comprises paired event data associated with the Caltech-101~\cite{fei2004learning} dataset. This dataset is employed for tasks encompassing event recognition, event-to-image retrieval, and image-to-event retrieval. The evaluation metric for the recognition task is accuracy, while for retrieval tasks, we utilize recall.

\noindent \textcolor{red!70!black}{\textbf{N-ImageNet-1K (N-IN-1K)}}~\cite{kim2021n}.
N-ImageNet-1K encompasses paired event data derived from the ImageNet-1K~\cite{deng2009imagenet} dataset. The evaluation focuses on assessing the event recognition capabilities of models within this dataset. Accuracy is employed as the metric for this evaluation.

\begin{figure*}[ht!]
    \centering
    \includegraphics[width=\textwidth]{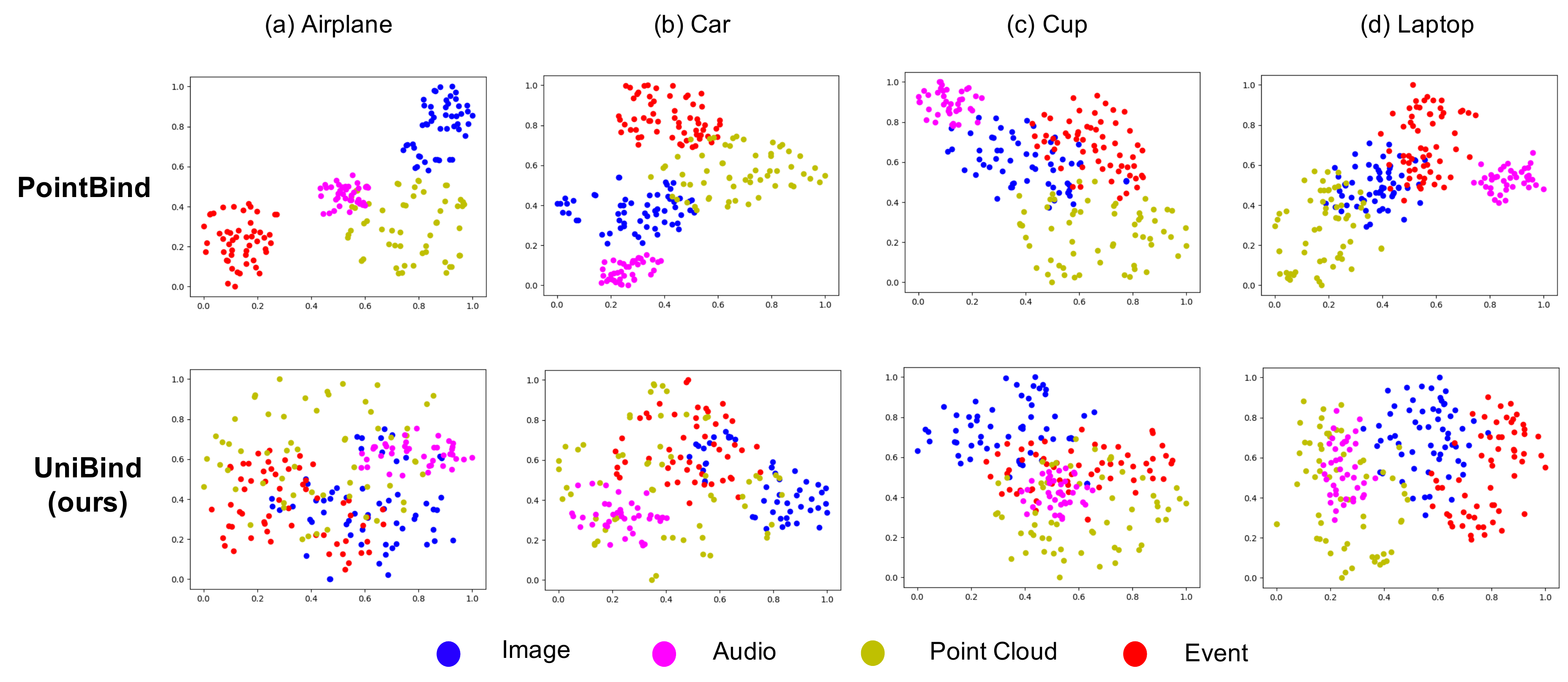}
    \caption{Representation space visualization of PointBind~\cite{guo2023point} and our UniBind.}
    \label{lal}
\end{figure*}

\begin{table*}[ht!]
\renewcommand{\tabcolsep}{10pt}
\resizebox{\linewidth}{!}{
\begin{tabular}{lcccccccc}
\toprule
\multicolumn{1}{c|}{Model} & \multicolumn{4}{c|}{\textcolor{blue!70!black}{Image}-to-\textcolor{red!70!black}{Event}} & \multicolumn{4}{c}{\textcolor{red!70!black}{Event}-to-\textcolor{blue!70!black}{Image}} \\ \cmidrule{2-9}
\multicolumn{1}{l|}{} & \multicolumn{1}{c|}{R@1} & \multicolumn{1}{c|}{R@5} & \multicolumn{1}{c|}{R@10} & \multicolumn{1}{c|}{R@20} & \multicolumn{1}{c|}{R@1} & \multicolumn{1}{c|}{R@5} & \multicolumn{1}{c|}{R@10} & \multicolumn{1}{c}{R@20} \\ \midrule
E-CLIP~\cite{zhou2023clip} & 79.52 & 90.11 & 93.08 & 95.51 & 76.29 & 89.97 & 91.80 & 94.61 \\
\textbf{E-CLIP w LCL}& 78.95 & 89.79 & 94.32 & 97.06 & 77.04 & 91.51 & 93.62 & 96.70 \\
\rowcolor{gray!10}$\Delta $ & -0.57 & -0.32 & \textbf{+1.24} & \textbf{+1.55} & \textbf{+0.75} & \textbf{+1.54} & \textbf{+1.82} & \textbf{+2.09}  \\ \midrule
PointBind (\textbf{+Event})~\cite{guo2023point} & 14.07 & 31.40 & 40.79 & 49.46 & 9.00 & 22.23 & 29.32 & 37.70  \\
\textbf{PointBind w LCL} & 14.12 & 31.91 & 41.25 & 50.98 & 14.29 & 33.65 & 44.34 & 55.66   \\
\rowcolor{gray!10}$\Delta $ & \textbf{+0.05} & \textbf{+0.51} & \textbf{+0.46} & \textbf{+1.52} & \textbf{+5.29} & \textbf{+11.42} & \textbf{+15.02} & \textbf{+17.96}   \\
\bottomrule
\end{tabular}}
\caption{Multi-modal retrieval result \textbf{with/without LLM-augmented contrastive Learning}. We evaluate E-CLIP and PointBind in Image-to-Event and Event-to-Image tasks.}
\vspace{-10pt}
\label{Table_ab_LG_2}
\end{table*}

\subsection{Experiment Details}

\subsubsection{Zero-shot Recognition}
In the zero-shot setting, we evaluate all baseline models and our UniBind without training. For all baseline models, we use the default templates from CLIP~\cite{radford2021learning}, and we use our localized embedding centers for our UniBind. 
\subsubsection{Fine-tuning Recognition}
For the fine-tuning setting, our experiments were done on 80GB A800 GPUs, and we detail the hyperparameters used for training with PointBind~\cite{guo2023point} reported in Tab~\ref{hypara}

\section{Additional Ablation Study}
\label{Additional Ablation Study}
\subsection{LLM-augmented Contrastive Learning}

We present additional visualization results are shown in Fig.~\ref{lal}. We show the comparison of the PointBind~\cite{guo2023point} representation space and our UniBind representation space. In the representation space built by PointBind, embeddings from different modalities tend to cluster around their respective modalities. Thereby, with LLM-augmented contrastive learning, multi-modal embeddings cluster around the same semantic label in our unified modality-agnostic representation space. 

We additionally present additional results pertaining to the cross-modal retrieval task. Our experimentation involves E-CLIP~\cite{zhou2023clip} and PointBind~\cite{guo2023point}, and we subsequently present the outcomes for both event-to-image retrieval and image-to-event retrieval in Table~\ref{Table_ab_LG_2}. The observed improvement in recall scores incrementally rises from the top 1 to the top 20, highlighting the effectiveness of our approach in aligning modalities with semantics.

\subsection{Embedding Center Localization}
\begin{figure*}[h!]
    \centering
    \includegraphics[width=\textwidth]{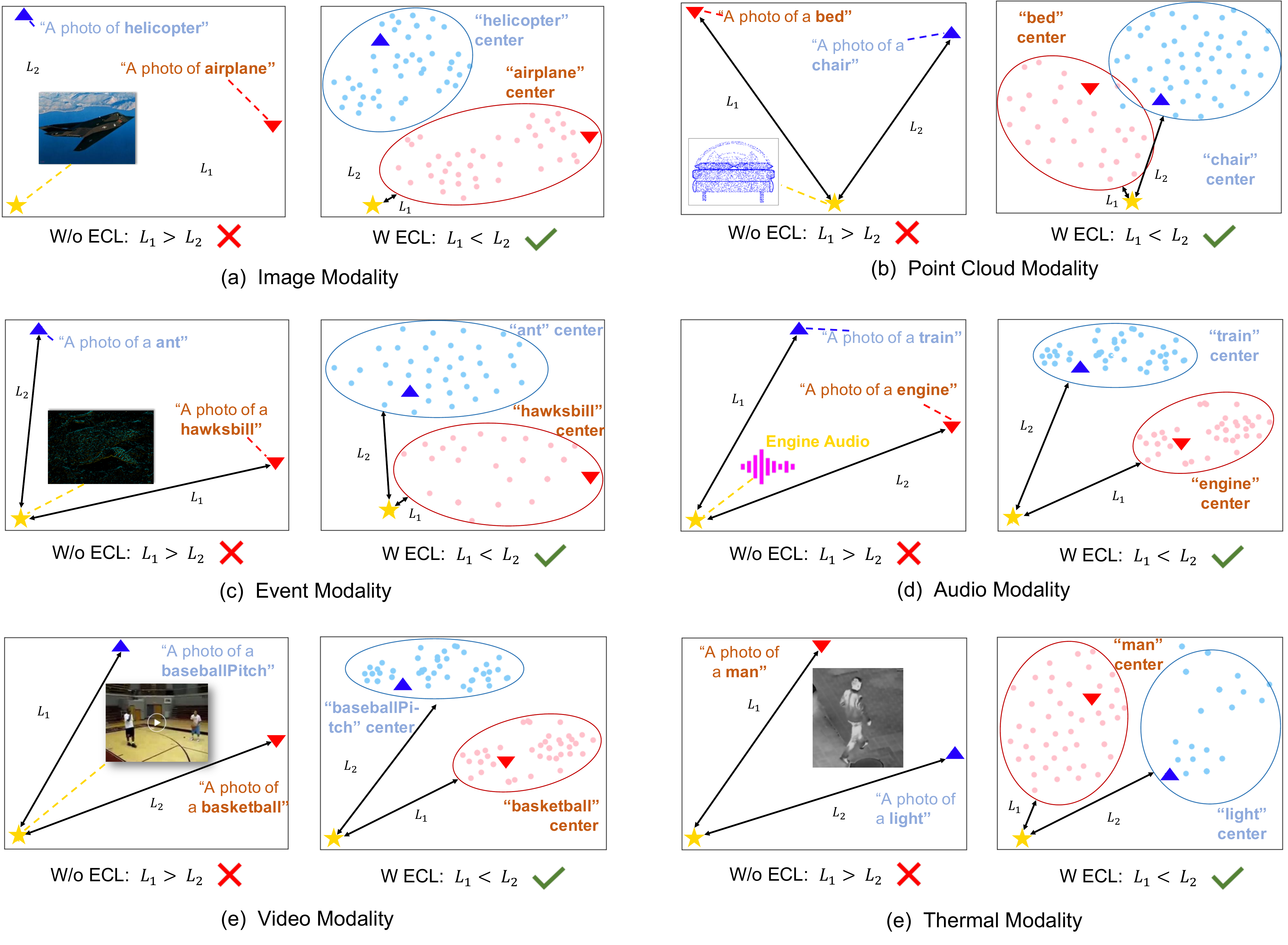}
    \caption{embedding centers.}
    \label{ECL}
\end{figure*}

We show more visualization results from image, point cloud, event, audio, video, and thermal modalities in Fig.~\ref{ECL}. Our embedding centers result in more distinct semantic boundaries between different categories, effectively enhancing recognition accuracy and reducing interference from other categories.

\end{document}